\documentclass[wcp]{jmlr}

\usepackage{longtable}

\usepackage{booktabs}

\usepackage{bm}
\usepackage{algorithmic}
\usepackage{algorithm}
\usepackage{subcaption}
\allowdisplaybreaks

\newtheorem{assumption}[theorem]{Assumption}

\makeatletter
\let\Ginclude@graphics\@org@Ginclude@graphics 
\renewcommand*{\@titlefoot}{}
\makeatother

\title[QHM using Increasing Batch Size]{Both Asymptotic and Non-Asymptotic Convergence of Quasi-Hyperbolic Momentum using Increasing Batch Size}

\author{Kento Imaizumi, Hideaki Iiduka}

\begin{document}

\maketitle

\begin{abstract}
Momentum methods were originally introduced for their superiority to stochastic gradient descent (SGD) in deterministic settings with convex objective functions. However, despite their widespread application to deep neural networks --- a representative case of stochastic nonconvex optimization --- the theoretical justification for their effectiveness in such settings remains limited. Quasi-hyperbolic momentum (QHM) is an algorithm that generalizes various momentum methods and has been studied to better understand the class of momentum-based algorithms as a whole. In this paper, we provide both asymptotic and non-asymptotic convergence results for mini-batch QHM with an increasing batch size. We show that achieving asymptotic convergence requires either a decaying learning rate or an increasing batch size. Since a decaying learning rate adversely affects non-asymptotic convergence, we demonstrate that using mini-batch QHM with an increasing batch size --- without decaying the learning rate --- can be a more effective strategy. Our experiments show that even a finite increase in batch size can provide benefits for training neural networks.
The code is available at \url{https://anonymous.4open.science/r/qhm_public}.
\end{abstract}
\begin{keywords}
nonconvex optimization; batch size; learning rate; momentum; asymptotic convergence; non-asymptotic convergence
\end{keywords}

\section{Introduction}
Stochastic gradient descent (SGD) \citep{10.1214/aoms/1177729586} is the simplest gradient method for minimizing empirical risk minimization problems. To improve its theoretical and empirical performance, many variants have been suggested, such as momentum methods and adaptive methods including Adam \citep{adam}, AdamW \citep{loshchilov2018decoupled}, and RMSProp \citep{rmsprop}. In particular, momentum methods are the most common variant and they accelerate convergence to optimal points by adding a momentum term, which is a combination of the current stochastic gradient. The initial goal of momentum methods is to speed up the convergence of convex optimization in the deterministic setting. In fact, the heavy-ball (HB) method \citep{polyak1964} and Nesterov’s accelerated gradient (NAG) \citep{Nesterov1983AMF} were shown to accelerate linear convergence over the standard gradient method under such conditions. On the other hand, several numerical experiments \citep{krizhevsky2012imagenet, 6296526, He2015DeepRL, 7780460, NIPS2016_577ef115, skipconv} employed stochastic variants of them --- stochastic HB (SHB) and stochastic NAG (SNAG) --- in deep learning for stochastic nonconvex optimization \citep{pmlr-v28-sutskever13}. Thus, there is a growing need for theoretical results demonstrating the usefulness of SHB and SNAG in environments that are the exact opposite of convex and deterministic settings.
 
\citep{ijcai2018p410} proved non-asymptotic convergence of stochastic unified momentum (SUM) (with an earlier preprint \citep{yang2016unifiedconvergenceanalysisstochastic}), including SGD, NSHB, and SNAG, for nonconvex optimization in the stochastic setting. Subsequently, \citep{pmlr-v97-yu19d} applied momentum to a distributed nonconvex setting. However, they imposed the condition $\alpha = \mathcal{O}\left(\frac{1}{\sqrt{K}}\right)$, where $\alpha$ is a constant learning rate and $K$ is the total number of steps, for which it is difficult to ensure asymptotic convergence and apply in practice. \citep{NEURIPS2020_d3f5d4de} gave an upper bound on the norm of the gradient of the objective function for NSHB $\min_{k\in [0:K-1]}\mathbb{E}[\|\nabla f(\bm{x}_k)\|^2] = \mathcal{O}\left(\frac{1}{K} + \sigma^2\right)$ in a nonconvex and stochastic environment without assuming the boundedness of the gradient, where $\nabla f:\mathbb{R}^d \to \mathbb{R}^d$ is the gradient of $f$, $\{\bm{x}_k\}_{k\in \mathbb{N}_0}$ is the sequence generated by NSHB, $\sigma^2$ is the upper bound of the variance of the stochastic gradient of $f$, and $\mathbb{E}[X]$ denotes the expectation of a random variable $X$. This analysis involved two major challenges. First, the learning rate $\alpha$ and momentum weight $\beta$ are constant, and this limits the practical learning rates of schedulers, such as cosine-annealing learning rate \citep{loshchilov2017sgdr} (Their proof is claimed to cover step decay learning rate, but it no longer applies if the learning rate depends on the iteration $k$). Second, due to the noise term, their analysis does not guarantee asymptotic convergence, i.e. $\liminf_{k \to \infty}\mathbb{E}[\|\nabla f(\bm{x}_k)\|] = \mathcal{O }(\sigma)$. 

In this paper, we focus on quasi-hyperbolic momentum (QHM) \citep{ma2018quasihyperbolic} (Section \ref{c_and_a}, Algorithm \ref{algo_qhm}). Similar to SUM, QHM becomes various optimizers depending on how $\gamma_k$ is set up, such as SGD ($\gamma_k=0$), Normalized-SHB (NSHB) \citep{nshb}($\gamma_k=1$), and SNAG ($\alpha_k$, $\beta_k$, and $\gamma_k$ are constant, and $\gamma_k$ = $\beta_k$) \citep{NEURIPS2019_4eff0720}. SHB and NSHB are interchangeable by adjusting the learning rate scheduler and momentum scheduler (see Appendix \ref{shb_and_nshb}). Hence, we believe that the analysis of QHM will contribute to our understanding of momentum methods. \citep{NEURIPS2019_4eff0720} proved asymptotic convergence of QHM under the condition of decreasing $\alpha_k$ and $\beta_k\gamma_k$ or under decreasing $\alpha_k$ and $\gamma_k$ and increasing $\beta_k$. 

The performance of optimizers used in a stochastic setting undoubtedly depends heavily on the batch size, and momentum methods are no exception \citep{10.1093/imanum/drae033, 10.1145/3447548.3467287}. Here, there are two main approaches to analyzing the batch size: small-batch learning and large-batch learning. In small-batch learning, the primary motivation is to identify the critical batch size that minimizes the number of stochastic first-order oracle (SFO) calls, which represent the computational cost of stochastic gradient evaluations \citep{Ghadimi:2016aa, pmlr-v202-sato23b, Imaizumi19062024}. This theory is based on the result that, although a larger batch size reduces the variance of stochastic gradients --- which in turn allows the noise term, a key challenge in non-asymptotic learning, to decrease linearly --- the number of steps required to reach an $\epsilon$--approximation does not continue to decrease linearly once the batch size exceeds a certain critical value \citep{shallue2019}. \citep{NIPS2017_a5e0ff62} indicates that the performance degradation of large-batch training is caused by an insufficient number of steps. In large-batch learning, the primary motivation is to speed up the training process by using the optimizer for large-batch training such as layer-wise adaptive rate scaling (LARS) \citep{you2017largebatchtrainingconvolutional} and layer-wise adaptive moments optimizer for batch training (LAMB) \citep{You2020Large}.

The above methods use a constant batch size $b$ that is independent of the iteration number $k$. \citep{l.2018dont} analyzed the existing results and found that decreasing the learning rate and increasing the batch size are equivalent in terms of regularization; they argued for increasing the batch size rather than decreasing the learning rate from the perspective of faster training. \citep{keskar2017on} pointed out two interesting insights about small batches and large batches from the perspective of sharpness of the loss function: (i) When using a large batch, the optimization tends to converge to sharp local minima, which leads to a significant drop in generalization performance. On the other hand, when using a small batch, the optimization avoids sharp local minima and tends to converge to flat local minima. (ii) Training with a small batch in the early stages and switching to a large batch in the later stages leads to convergence to flatter local minima compared with using only a small batch. Although both of these studies claim the benefit of an increasing batch size, they lack theoretical support.

\subsection{Contribution}
In this paper, we focus on mini-batch QHM (QHM using a batch size) with various learning rate and momentum weight schedulers and prove both asymptotic and non-asymptotic convergence of nonconvex optimization in the stochastic setting.
\begin{itemize}
\item To compare constant and increasing batch-size settings, we provide asymptotic and non-asymptotic convergence analyses of mini-batch QHM, NSHB, and SGD in stochastic nonconvex optimization.
\end{itemize}

Specifically, we find that mini-batch QHM, NSHB, and SGD achieve asymptotic convergence by setting $\alpha_k \to +\infty$, $\beta_k\gamma_k \to 0$, and $\frac{\alpha_k^2}{b_k} \to 0$, where $b_k$ is the batch size. This enables the use of practical learning-rate schedules such as a constant learning rate, cosine annealing, and polynomial decay, which would otherwise require a decaying learning rate under a constant batch size. In the analysis of non-asymptotic convergence, mini-batch QHM, NSHB, and SGD using a constant learning rate, step decay momentum weights, and constant batch size are shown to have a convergence rate of $\mathcal{O}\left(\dfrac{1}{K} + \dfrac{\sigma^2}{b}\right)$. These results suggest that the inability to achieve asymptotic convergence is due to the noise term. While using a large batch size can reduce the upper bound, it does not guarantee convergence. On the other hand, in the case of using an increasing batch size \citep{pmlr-v54-de17a, l.2018dont, NEURIPS2024_ef74413c, umeda2025increasing, Sato_Iiduka_2025}, the convergence rate improves to $\mathcal{O}\left(\dfrac{1}{K}\right)$, overcoming the challenge of the noise term.

\begin{itemize}
\item While a decaying learning rate has the advantage of achieving asymptotic convergence when using a constant batch size, it also becomes a cause of performance degradation in non-asymptotic convergence.
\end{itemize}

In the case of using an increasing batch size, mini-batch QHM, NSHB, and SGD using a decaying learning rate and step decay momentum weights have a convergence rate of $\mathcal{O}\left(\dfrac{1}{\sqrt{K}}\right)$, i.e., slower convergence compared with using practical learning rates, including constant learning rates. We also provide asymptotic and non-asymptotic convergence results under different conditions to allow $\gamma_k$ to be fixed to an arbitrary constant, although this requires a decaying learning rate. The case of a constant batch size requires the condition $\beta_k \to 1$, which further requires the learning rate to decay faster than the increase in $\beta_k$, leading to degraded performance. In the case of an increasing batch size, the condition is relaxed so that $\beta_k$ can be fixed, but since a decaying learning rate is still used, performance remains suboptimal.

\begin{itemize}
\item We empirically compared cases of using an increasing batch size and using a small batch size or large batch size in training deep neural networks with mini-batch QHM, NSHB, and SGD under practical learning rates, including a decaying learning rate. The results support the theoretical claims.
\end{itemize}

Our theoretical condition $b_k \to +\infty$ is a rather strong assumption. However, empirical results show that we can still benefit from increasing batch sizes, even if the upper bound grows.

\section{Preliminaries}
\subsection{Notation}
Let $\mathbb{R}^d$ be a $d$-dimensional Euclidean space with inner product $\langle \bm{x},\bm{y} \rangle := \bm{x}^\top \bm{y}$ ($\bm{x},\bm{y} \in \mathbb{R}^d$) inducing the norm $\| \bm{x}\| := \sqrt{\langle \bm{x}, \bm{x}} \rangle$, and let $\mathbb{N}$ and $\mathbb{N}_0$ be the set of natural numbers and the set of natural numbers including zero, respectively. Define $[n] := \{1,2,\ldots,n\}$ and $[0:n] := \{0,1,\ldots,n\}$ for $n \geq 1$. Let $\{x_k\}_{k\in\mathbb{N}}$ and $\{y_k\}_{k\in\mathbb{N}}$ be positive real sequences and let $x(\epsilon), y(\epsilon) > 0$, where $\epsilon > 0$. $\mathcal{O}$ denotes Landau's symbol; i.e., $y_k = \mathcal{O}(x_k)$ if there exist $c > 0$ and $k_0 \in \mathbb{N}$ such that $y_k \leq c x_k$ for all $k \geq k_0$, and $y(\epsilon) = \mathcal{O} (x(\epsilon))$ if there exists $c > 0$ such that $y(\epsilon) \leq cx(\epsilon)$. Given a parameter $\bm{x} \in \mathbb{R}^d$ and a data point $z$ in a data domain $Z$, a machine-learning model provides a prediction whose quality can be measured by a differentiable nonconvex loss function $\ell(\bm{x};z)$. We aim to minimize the empirical loss defined for all $\bm{x} \in \mathbb{R}^d$ by $f(\bm{x}) = \frac{1}{n} \sum_{i=1}^n \ell(\bm{x};z_i) = \frac{1}{n} \sum_{i=1}^n f_i(\bm{x})$, where $S = \{z_1, z_2, \ldots, z_n\}$ denotes the training set and $f_i (\cdot) := \ell(\cdot;z_i)$ denotes the loss function corresponding to the $i$-th training data $z_i$.

\subsection{Conditions and Algorithm} 
\label{c_and_a}
The following are standard conditions.
\begin{enumerate}
\item[(C1)]$f := \frac{1}{n} \sum_{i=1}^n f_i \colon \mathbb{R}^d \to \mathbb{R}$ is $L$--smooth, i.e. $\|\nabla f (\bm{x}) - \nabla f(\bm{y})\| \leq L \|\bm{x} - \bm{y} \|$. $f$ is bounded below from $f_\star \in \mathbb{R}$.
\item[(C2)]Let $(\bm{x}_k)_{k\in \mathbb{N}} \subset \mathbb{R}^d$ be the sequence generated by mini-batch QHM. For each iteration $k$, $\mathbb{E}_{\xi_k} [ \nabla f_{\xi_k}(\bm{x}_k) ] = \nabla f(\bm{x}_k)$, where $\xi_0, \xi_1, \ldots$ are independent samples and the random variable $\xi_k$ is independent of $(\bm{x}_l)_{l=0}^k$. There exists a nonnegative constant $\sigma^2$ such that $\mathbb{V}_{\xi_k} [ \nabla f_{\xi_k}(\bm{x}_k)] \leq \sigma^2$, where $\mathbb{V}[X]$ denotes the variance of a random variable $X$.
\item[(C3)] For each iteration $k$, mini-batch QMH samples a batch $B_{k}$ of size $b_k$ independently of $k$ and estimates the full gradient $\nabla f$ as $\nabla f_{B_k} (\bm{x}_k) := \frac{1}{b_k} \sum_{i\in [b_k]} \nabla f_{\xi_{k,i}}(\bm{x}_k)$, where $\xi_{k,i}$ is a random variable generated by the $i$-th sampling in the $k$-th iteration. 
\end{enumerate}

Algorithm \ref{algo_qhm} is the mini-batch QHM optimizer under (C1) -- (C3).

\begin{algorithm} 
\caption{Mini-batch Quasi-Hyperbolic Momentum (mini-batch QHM)} 
\label{algo_qhm} 
\begin{algorithmic}[1] 
\REQUIRE
$\bm{x}_{0} \in\mathbb{R}^d$ (initial point), $\alpha_k \in (0,+\infty)$ (learning rate), $\beta_k \in [0, 1), \gamma_k \in [0,1]$ (momentum weights), $b_k \in \mathbb{N}$ (batch size), $K \in \mathbb{N}$ (steps)
\ENSURE
$\bm{x}_{K}$
\FOR{$k = 0, 1, \ldots, K-1$}
\STATE
$\nabla f_{B_k} (\bm{x}_k)
:= \frac{1}{b_k} \sum_{i\in [b_k]} \nabla f_{\xi_{k,i}}(\bm{x}_k)$
\STATE
$\bm{d}_k := (1 - \beta_k)\nabla f_{B_k}(\bm{x}_k) + \beta_k\bm{d}_{k-1}$
\STATE
$\bm{m}_k := (1 - \gamma_k)\nabla f_{B_k}(\bm{x}_k) + \gamma_k\bm{d}_k$
\STATE
$\bm{x}_{k+1} := \bm{x}_k - \alpha_k\bm{m}_k$
\ENDFOR 
\end{algorithmic}
\end{algorithm}

\subsection{Batch Size Schedulers}
Let $M \in \mathbb{N}$ be the total number of epochs, and let $E \in \mathbb{N}$ be the number of interval epochs. The batch size is updated per epoch, not per step. So, the batch size schedulers below are are indexed by the number of epochs $m \in [0:M-1]$:
\begin{align}
&\text{[Constant BS] } b'_{m} = b, \label{constant_batch} \\
&\text{[Exponential BS] } b'_{m} = b_0\delta^{\left\lfloor\frac{m}{E}\right\rfloor}, \label{exp_batch}
\end{align}
where $b, b_0 \in \mathbb{N}$ and $\delta > 1$. $b'_{m}$ is related to the batch size indexed by the step $k$, as follows:
\begin{align*}
&\{b_k\}_{k \in \mathbb{N}_0}
= 
\{\underbrace{b'_0,\ldots, b'_0}_{T_0}, \underbrace{b'_1,\ldots, b'_1}_{T_1},\ldots,\underbrace{b'_{M-1},\ldots, b'_{M-1}}_{T_{M-1}},\ldots\},
\end{align*}
where $T_m = \left\lceil\dfrac{n}{b_m'}\right\rceil$ is the number of steps per epoch $m$.

\subsection{Learning Rate Schedulers}
We will examine the following learning rate schedules that are updated each epoch $m \in [0:M-1]$:
\begin{align}
&\text{[Constant LR] } {\alpha'}_m = \alpha_{\max}, \label{constant_lr} \\
&\text{[Decaying Squared LR] } {\alpha'}_m = \frac{\alpha_{\max}}{\sqrt{m + 1}}, \label{decaying_lr_sq} \\
&\text{[Decaying LR] } {\alpha'}_m = \frac{\alpha_{\max}}{m + 1}, \label{decaying_lr} \\
&\text{[Cosine-annealing LR] } {\alpha'}_m = \alpha_{\min} + \frac{\alpha_{\max} - \alpha_{\min}}{2}\left(1 + \cos\frac{m\pi}{M-1}\right), \label{cosine_lr} \\
&\text{[Polynomial Decay LR] } {\alpha'}_m = \alpha_{\min} + (\alpha_{\max} - \alpha_{\min})\left(1 - \frac{m}{M}\right)^p, \label{poly_lr}
\end{align}
where $p > 0$, and $0 \leq \alpha_{\min} \leq \alpha_{\max}$. In the case of [Cosine LR], ${\alpha'}_m = \alpha_{\max}$ if $M = 1$, whereas in the case of [Cosine LR] or [Polynomial LR], the total number of steps $K \in \mathbb{N}$ satisfies $K = \sum_{m=0}^{M-1}T_m$. ${\alpha'}_{m}$ is related to the learning rate indexed by the step $k$ in the same way as the batch size.

\subsection{Momentum weight schedulers}
We will examine the following schedules for the momentum weights $\gamma_k$ and $\beta_k$ in Algorithm \ref{algo_qhm} that are updated per epoch:
\begin{align}
&\text{[Constant Beta] }
\beta'_m = \beta_{\max}, \label{constant_beta} \\
&\text{[Step Decay Beta] }
\beta'_m = \beta_{\max}\zeta^{\left\lfloor\frac{m}{E}\right\rfloor}, \label{step_decay_beta} \\
&\text{[Increasing Beta] }
\beta'_m = 1 - \frac{1 - \beta_{\min}}{(m + 1)^{3/4}}, \label{inc_beta} \\
&\text{[Constant Gamma] }
\gamma'_m = \gamma_{\max}, \label{constant_gamma} \\
&\text{[Step Decay Gamma] }
\gamma'_m = \gamma_{\max}\lambda^{\left\lfloor\frac{m}{E}\right\rfloor}. \label{step_decay_gamma}
\end{align}
Here, $\lambda, \zeta > 0$, $\beta_{\min}, \beta_{\max} \in [0,1)$, and $\gamma_{\max} \in [0,1]$. $\beta'_k$ and $\gamma'_k$ are related to the momentum weights indexed by the step $k$ in the same way as the batch size.

The reason for updating the scheduler per epoch rather than per step is that the principle of "constant and drop", where the learning rate is large initially and then dropped every few epochs, is well-known for adjusting the learning rate. Algorithms that apply the idea of "constant and drop" are referred to as multistage algorithms \citep{NEURIPS2020_d3f5d4de, 10.1145/3447548.3467287}.

\section{Asymptotic and Non-Asymptotic Convergence of Quasi-Hyperbolic Momentum}
In this section, we show both asymptotic and non-asymptotic convergence of QHM with various momentum weights schedulers and argue that an increasing batch size schedule should be used rather than a constant batch size.

In order to prove the following theorems, we will make an additional assumption that is the same as the one made in \citep{ijcai2018p410} and \citep{NEURIPS2019_4eff0720}.

\begin{assumption}
[Boundedness of the gradient] \label{b_grad} Let $G > 0$ satisfy, for all $i \in [n]$ and $\bm{x} \in \mathbb{R}^d$, 
\begin{align*}
\| \nabla f_i(\bm{x}) \| \leq G.
\end{align*}
\end{assumption}

\begin{theorem}
[Upper bound of the squared norm of the full gradient using decreasing $\beta_k\gamma_k$] \label{th_1}
Suppose that conditions (C1)--(C3) and Assumption \ref{b_grad} hold, and let $\alpha_k \in [\alpha_{\min}, \alpha_{\max}] \subset [0, \frac{2}{L})$, $\overline{\gamma\beta} := \sup_k\gamma_k\beta_k < 1$. Then, the sequence $\{\bm{x}_k\}_{k\in \mathbb{N}_0}$ generated by Algorithm \ref{algo_qhm} has the following properties.

\noindent {\em (i)} {\em [Convergence Rate]} For all $K \in \mathbb{N}$,
\begin{align*}
\min_{k \in [0:K-1]}\mathbb{E}\left[\left\|\nabla f(\bm{x}_k)\right\|^2\right] &\leq
\frac{2(f(\bm{x}_0) - f_\star)}{(1-\overline{\gamma\beta})(2-\alpha_{\max}L)}\underbrace{\frac{1}{\sum_{k=0}^{K-1}\alpha_k}}_{A_K} + \frac{L\sigma^2}{(1-\overline{\gamma\beta})(2-\alpha_{\max}L)}\underbrace{\frac{\sum_{k=0}^{K-1}\alpha_k^2/b_k}{\sum_{k=0}^{K-1}\alpha_k}}_{B_K} \\
&\quad + \frac{2\alpha_{\max}(1+\alpha_{\max}L)G^2}{(1-\overline{\gamma\beta})(2-\alpha_{\max}L)}\underbrace{\frac{\sum_{k=0}^{K-1}\gamma_k\beta_k}{\sum_{k=0}^{K-1}\alpha_k}}_{C_K}.
\end{align*}
{\em (ii)} {\em [Convergence]} Additionally, let $\{\alpha_k\}$, $\{\beta_k\}$, $\{\gamma_k\}$, and $\{b_k\}$ satisfy the following conditions:
\begin{align*}
&\sum_{k=0}^{+\infty}\alpha_k = +\infty, \quad \sum_{k=0}^{+\infty}\gamma_k\beta_k < +\infty, \quad \sum_{k=0}^{+\infty}\frac{\alpha_k^2}{b_k} < +\infty.
\end{align*}
Then,
\begin{align*}
\liminf_{k \to \infty} \mathbb{E}\left[\| \nabla f(\bm{x}_k) \| \right] = 0.
\end{align*}
\end{theorem}

Theorem \ref{th_1} asserts that, for QHM with decreasing $\gamma_k\beta_k$, achieving non-asymptotic convergence requires either a decreasing learning rate or increasing batch size, and it is based on the result of \citep{l.2018dont} on empirical risk minimization problems. In the case of using [Cosine LR] or [Polynomial LR], $\alpha_k$ is only defined up to $K = \sum_{m=0}^{M-1}T_m$. Hence, we employ the following learning-rate schedulers for Theorem \ref{th_1}(ii):
\begin{align*}
\alpha_k =
\begin{cases}
\text{[Cosine LR \eqref{cosine_lr}] or [Polynomial Decay \eqref{poly_lr}]} & \left(k \leq \sum_{m=0}^{M-1}T_m\right) \\
\text{[Decaying Squared LR \eqref{decaying_lr_sq}]} & \left(k > \sum_{m=0}^{M-1}T_m\right).
\end{cases}
\end{align*}

Note that this is introduced for the theoretical analysis and not in the numerical experiments or Theorem \ref{th_1}(i).

The following Corollary \ref{cor_1} considers non-asymptotic convergence in the case of using [Step Decay Beta \eqref{step_decay_beta}] and [Step Decay Gamma \eqref{step_decay_beta}] in accordance with Theorem \ref{th_1}, and it asserts that QHM using [Constant BS \eqref{constant_batch}] and a learning rate with upper and lower bounds, such as [Constant LR \eqref{constant_lr}], [Cosine LR \eqref{cosine_lr}], and [Polynomial LR \eqref{poly_lr}], does not converge to a local minimum because $B_K$ does not converge to $0$. While using [Decaying Squared LR \eqref{decaying_lr_sq}] can address the issue of the noise term even with [Constant BS \eqref{constant_batch}], it slows convergence.

\begin{corollary}
[Upper bound of $A_K$, $B_K$, and $C_K$ in Theorem \ref{th_1}] \label{cor_1} When using [Step Decay Beta \eqref{step_decay_beta}] and [Step Decay Gamma \eqref{step_decay_gamma}], $A_K$, $B_K$, and $C_K$ in Theorem \ref{th_1} satisfy, for all $K \in \mathbb{N}$,
\begin{align*}
&A_K \leq
\begin{cases}
\frac{1}{\alpha_{\max}K} &[\text{Constant LR \eqref{constant_lr}}] \\
\frac{1}{2\alpha_{\max}\left(\sqrt{K+1}-1\right)} &[\text{Decaying Squared LR \eqref{decaying_lr_sq}}] \\
\frac{2}{(\alpha_{\min} + \alpha_{\max})K} &[\text{Cosine LR \eqref{cosine_lr}}] \\
\frac{p+1}{(\alpha_{\max} + p\alpha_{\min})K}&[\text{Polynomial LR \eqref{poly_lr}}],
\end{cases} \\
&B_K \leq
\begin{cases}
\frac{\alpha_{\max}}{b} &[\text{Constant LR \eqref{constant_lr}}] \\
\frac{\alpha_{\max}T_0\left(1+\log(K+1)\right)}{2b(\sqrt{K+1}-1)} &[\text{Decaying Squared LR \eqref{decaying_lr_sq}}] \\
\frac{\alpha_{\max}+\alpha_{\min}}{b}+\frac{T_0(\alpha_{\max}-\alpha_{\min})^2}{2b(\alpha_{\max}+\alpha_{\min})K} &[\text{Cosine LR \eqref{cosine_lr}}] \\
\frac{(p+1)\alpha_{\max}^2+2p\alpha_{\max}\alpha_{\min}+2p^2\alpha_{\min}^2}{b(2p+1)(\alpha_{\max}+p\alpha_{\min})} + \frac{(p+1)(\alpha_{\max}^2-\alpha_{\min}^2)T_0}{(\alpha_{\max}+p\alpha_{\min})K} &[\text{Polynomial LR \eqref{poly_lr}}]
\end{cases}
[\text{Constant LR \eqref{constant_lr}}], \\
&B_K \leq
\begin{cases}
\frac{\alpha_{\max}T_0E\delta}{b_0(\delta-1)K} &[\text{Constant LR \eqref{constant_lr}}] \\
\frac{\alpha_{\max}T_0E\delta}{2b_0(\delta-1)(\sqrt{K+1}-1)} &[\text{Decaying Squared LR \eqref{decaying_lr_sq}}] \\
\frac{2\alpha_{\max}^2T_0E\delta}{b_0(\delta-1)(\alpha_{\max}+\alpha_{\min})K} &[\text{Cosine LR} \eqref{cosine_lr}] \\
\frac{\alpha_{\max}^2(p+1)T_{0}E\delta}{b_0(\alpha_{\max}+p\alpha_{\min})(\delta-1)K} &[\text{Polynomial LR} \eqref{poly_lr}]
\end{cases}
[\text{Exponential BS \eqref{exp_batch}}], \\
&C_K \leq
\begin{cases}
\frac{\gamma_{\max}\beta_{\max}T_0E}{\alpha_{\max}(1-\lambda\zeta)K} &[\text{Constant LR \eqref{constant_lr}}] \\
\frac{\gamma_{\max}\beta_{\max}T_0E}{2\alpha_{\max}(1-\lambda\zeta)(\sqrt{K+1}-1)} &[\text{Decaying Squared LR \eqref{decaying_lr_sq}}] \\
\frac{\gamma_{\max}\beta_{\max}T_0E}{(1-\lambda\zeta)(\alpha_{\max}+\alpha_{\min})K} &[\text{Cosine LR \eqref{cosine_lr}}] \\
\frac{\gamma_{\max}\beta_{\max}(p+1)T_0E}{(\alpha_{\max}+p\alpha_{\min})(1-\lambda\zeta)K} &[\text{Polynomial LR \eqref{poly_lr}}],
\end{cases}
\end{align*}

That is, Algorithm \ref{algo_qhm} has the following convergence rates: in the case of using [\text{Constant BS \eqref{constant_batch}}],
\begin{align*}
\min_{k \in [0:K-1]}\mathbb{E}\left[\left\|\nabla f(\bm{x}_k)\right\|^2\right] \leq
\begin{cases}
\mathcal{O}\left(\frac{1}{K} + \frac{\sigma^2}{b}\right) & [\text{Constant LR} \eqref{constant_lr}] [\text{Cosine LR} \eqref{cosine_lr}][\text{Polynomial LR} \eqref{poly_lr}] \\
\mathcal{O}\left(\frac{\log K}{\sqrt{K}}\right) & [\text{Decaying Squared LR} \eqref{decaying_lr_sq}],
\end{cases}
\end{align*}
and in the case of using [\text{Exponential BS \eqref{exp_batch}}],
\begin{align*}
\min_{k \in [0:K-1]}\mathbb{E}\left[\left\|\nabla f(\bm{x}_k)\right\|^2\right] \leq
\begin{cases}
\mathcal{O}\left(\frac{1}{K}\right) & [\text{Constant LR} \eqref{constant_lr}][\text{Cosine LR} \eqref{cosine_lr}][\text{Polynomial LR} \eqref{poly_lr}] \\
\mathcal{O}\left(\frac{1}{\sqrt{K}}\right) & [\text{Decaying Squared LR} \eqref{decaying_lr_sq}].
\end{cases}
\end{align*}

\end{corollary}

Theorem \ref{th_1} has limited application to NSHB under the conditions $\gamma_k=1$ and either decreasing $\beta_k$ or decreasing $\gamma_k$. In particular, \citep{pmlr-v28-sutskever13} uses SNAG with an increasing $\beta_k$. Thus, Theorem \ref{th_2} assumes an increasing $\beta_k$.

\begin{theorem}
[Upper bound of the squared norm of the full gradient using increasing $\beta_k$] \label{th_2} Suppose that conditions (C1)--(C3) and Assumption \ref{b_grad} hold and that $\alpha_k$ and $\beta_k$ satisfy, for all $k \in \mathbb{N}_0$,
\begin{align*}
\beta_k\left(\frac{\alpha_k}{2} + 1\right) \leq 1.
\end{align*}
Then, the sequence $\{\bm{x}_k\}_{k\in \mathbb{N}_0}$ generated by Algorithm \ref{algo_qhm}has the following properties:

\noindent {\em (i)} {\em [Convergence Rate]} For all $K \in \mathbb{N}$,

\begin{align*}
\min_{k \in [0:K-1]}\mathbb{E}\left[\left\|\nabla f(\bm{x}_k)\right\|^2\right] &\leq 2(f(\bm{x}_0) - f_\star)\underbrace{\frac{1}{\sum_{k=0}^{K-1}\alpha_k}}_{A_K} + 4\sigma^2\underbrace{\sum_{k=0}^{K-1}\frac{1-\beta_k}{b_k}\frac{1}{\sum_{k=0}^{K-1}\alpha_k}}_{B'_k} \\
&\quad + 5L^2G^2\underbrace{\frac{\sum_{k=0}^{K-1}\alpha_k^2}{\sum_{k=0}^{K-1}\alpha_k}}_{C'_K} + 2L^2G^2\underbrace{\sum_{k=0}^{K-1}\frac{\alpha_k^2}{1-\beta_k}\frac{1}{\sum_{k=0}^{K-1}\alpha_k}}_{D'_K}.
\end{align*}
{\em (ii)} {\em [Convergence]} Additionally, let $\{\alpha_k\}$ and $\{\beta_k\}$ satisfy the following conditions:
\begin{align*}
&\sum_{k=0}^{+\infty}\alpha_k = +\infty, \quad \sum_{k=0}^{+\infty}\frac{\alpha_k^2}{1-\beta_k} < +\infty, \quad \sum_{k=0}^{+\infty}\frac{1-\beta_k}{b_k} < +\infty.
\end{align*}
Then,
\begin{align*}
\liminf_{k \to \infty} \mathbb{E}\left[\| \nabla f(\bm{x}_k) \| \right] = 0.
\end{align*}
\end{theorem}

Theorem \ref{th_2} indicates that, for QHM with decreasing $\alpha_k$, achieving non-asymptotic convergence requires either increasing $\beta_k$ and decreasing $\alpha_k^2$ faster than $\beta_k$ is increased or increasing the batch size and that it does not depend on $\gamma_k$. Hence, we can take $\gamma_k$ to be a constant.

The condition $\beta_k(\frac{\alpha_k}{2} + 1) \leq 1$ is practical because $\beta_{\max} \leq 0.9523 \cdots$ should be chosen if $\alpha_{\max} = 0.1$ is assumed and $\alpha_{\max} \leq 0.1$ should be chosen if $\beta_{\min} = 0.5$. As in Corollary \ref{cor_1}, Corollary \ref{cor_2} below considers non-asymptotic convergence in the case of using [Decaying LR \eqref{decaying_lr}] or [Decaying Squared LR \eqref{decaying_lr_sq}] in accordance with Theorem \eqref{th_2}; it asserts that QHM using [Constant BS \eqref{constant_batch}] and [Constant Beta \eqref{constant_beta}] does not converge to a local minimum because $B_K$ does not converge to $0$.

\begin{corollary}
[Upper bound of $B'_K$, $C'_K$, and $D'_K$ in Theorem \ref{th_2}] \label{cor_2} When using [Decaying LR \eqref{decaying_lr}] or [Decaying Squared LR \eqref{decaying_lr_sq}], $B'_K$, $C'_K$, and $D'_K$ in Theorem \ref{th_2} satisfy, for all $K \in \mathbb{N}$,

\begin{align*}
&A'_K \leq
\begin{cases}
\frac{1}{2\alpha_{\max}\left(\sqrt{K+1}-1\right)} &[\text{Decaying Squared LR \eqref{decaying_lr_sq}}] \land [\text{Constant Beta \eqref{constant_beta}}]\\
\frac{1}{\alpha_{\max}\log(K+1)} &[\text{Decaying LR \eqref{decaying_lr}}] \land [\text{Increasing Beta \eqref{inc_beta}}],
\end{cases} \\
&B'_K \leq
\begin{cases}
\begin{cases}
\frac{K(1-\beta_{\max})}{b(\sqrt{K+1}-1)} &[\text{Decaying Squared LR \eqref{decaying_lr_sq}}] \land [\text{Constant Beta \eqref{constant_beta}}]\\
\frac{T_0\left(1+4(1-\beta_{\min})(K+1)^{1/4}\right)}{b\alpha_{\max}\log(K+1)} &[\text{Decaying LR \eqref{decaying_lr}}] \land [\text{Increasing Beta \eqref{inc_beta}}],
\end{cases} 
[\text{Constant BS \eqref{constant_batch}}] \\
\begin{cases}
\frac{(1-\beta_{\max})T_0E\delta}{2b_0\alpha_{\max}(\delta-1)(\sqrt{K+1}-1)} &[\text{Decaying Squared LR \eqref{decaying_lr_sq}}] \land [\text{Constant Beta \eqref{constant_beta}}] \\
\frac{T_0E\delta}{b_0\alpha_{\max}(\delta-1)\log(K+1)} &[\text{Decaying LR \eqref{decaying_lr}}] \land [\text{Increasing Beta \eqref{inc_beta}}]
\end{cases}
[\text{Exponential BS \eqref{exp_batch}}],
\end{cases} \\
&C'_K \leq
\begin{cases}
\frac{\alpha_{\max}T_0\left(1+\log(K+1)\right)}{2(\sqrt{K+1}-1)} &[\text{Decaying Squared LR \eqref{decaying_lr_sq}}] \\
\frac{2\alpha_{\max}T_0}{\log(K+1)} &[\text{Decaying LR \eqref{decaying_lr}}],
\end{cases} \\
&D'_K \leq
\begin{cases}
\dfrac{\alpha_{\max }}{(1-\beta_{\max})(\sqrt{K+1}-1)} &[\text{Decaying Squared LR \eqref{decaying_lr_sq}}] \land [\text{Constant Beta \eqref{constant_beta}}] \\
\dfrac{2\alpha_{\max}T_0}{(1-\beta_{\min})\log(K+1)} &[\text{Decaying LR \eqref{decaying_lr}}] \land [\text{Increasing Beta \eqref{inc_beta}}].
\end{cases}
\end{align*}

That is, Algorithm \ref{algo_qhm} has the following convergence rates: in the case of using [Constant BS \eqref{constant_batch}],
\begin{align*}
\min_{k \in [0:K-1]}\mathbb{E}\left[\left\|\nabla f(\bm{x}_k)\right\|^2\right] \leq
\begin{cases}
\mathcal{O}\left({\sqrt{K}}\right) &[\text{Decaying Squared LR \eqref{decaying_lr_sq}}] \land [\text{Constant Beta \eqref{constant_beta}}] \\
\mathcal{O}\left(\frac{1}{\log K}\right) &[\text{Decaying LR \eqref{decaying_lr}}] \land [\text{Increasing Beta \eqref{inc_beta}}], \\
\end{cases}
\end{align*}
and in the case of [Exponential BS \eqref{exp_batch}],
\begin{align*}
\min_{k \in [0:K-1]}\mathbb{E}\left[\left\|\nabla f(\bm{x}_k)\right\|^2\right] \leq
\begin{cases}
\mathcal{O}\left({\frac{\log K}{\sqrt{K}}}\right) &[\text{Decaying Squared LR \eqref{decaying_lr_sq}}] \land [\text{Constant Beta \eqref{constant_beta}}] \\
\mathcal{O}\left(\frac{1}{\log K}\right) &[\text{Decaying LR \eqref{decaying_lr}}] \land [\text{Increasing Beta \eqref{inc_beta}}]. \\
\end{cases}
\end{align*}
\end{corollary}

Because [Decaying LR \eqref{decaying_lr}] decreases more rapidly than [Decaying Squared LR \eqref{decaying_lr_sq}], the non-asymptotic convergence performance is further degraded. While the use of [Decaying Squared LR \eqref{decaying_lr_sq}] and [Constant Beta \eqref{constant_beta}] leads to divergence in the case of a fixed batch size, in practice, the performance of the setting described in Theorem \ref{th_1} --- employing [Decaying Squared LR \eqref{decaying_lr_sq}], [Constant Beta \eqref{constant_beta}], and [Constant Gamma \eqref{constant_gamma}] --- is characterized by $\mathcal{O}(\frac{\log K}{\sqrt{K}})$.

\section{Numerical Results}
We experimented with mini-batch QHM and NSHB using various lr and momentum weight schedulers to show that they benefit from using an increasing batch size. Specifically, we examined the test accuracy and the minimum of the full gradient norm of the empirical loss in training ResNet--18 \citep{He2015DeepRL} on the CIFAR--100 dataset \citep{Krizhevsky2009LearningML} by using a small batch size ($b = 2^{8}$), large batch size ($b = 2^{12}$), and increasing batch size. Our experiments demonstrated that the scheduler with the increasing batch size is preferable, as neither the small nor large batch size proved effective. The experimental environment was an NVIDIA A100 PCIe 80GB GPU. The software environment was Python 3.11.11, Pytorch 2.2.2, and CUDA 12.2. The code is available at \url{https://anonymous.4open.science/r/qhm_public}.

We set the total number of epochs to $M=300$, the interval epochs to $E = 30$, the common ratios in \eqref{step_decay_beta} and \eqref{step_decay_gamma} to $\lambda = \zeta = 2$, the minimum learning rates in \eqref{cosine_lr} and \eqref{poly_lr} to $\alpha_{\min} = 0$, and the power in \eqref{poly_lr} to $p=2$. The other parameters in the lr and momentum weight schedulers were determined by a grid search to be $\alpha_{\max} \in \{0.05, 0.1, 0.25, 0.5\}$, $\beta_{\max}, \beta_{\min} \in \{0.3, 0.5, 0.9\}$, and $\gamma_{\max} \in \{0.2, 0.4, 0.7\}$.

The numerical results reported below are average results over three runs with different random seeds for the initial values. The range between the maximum and minimum values is illustrated using a shaded region.

\begin{figure}[htbp]
\centering
\subfigure[Constant LR]{
	\includegraphics[width=0.441\linewidth]{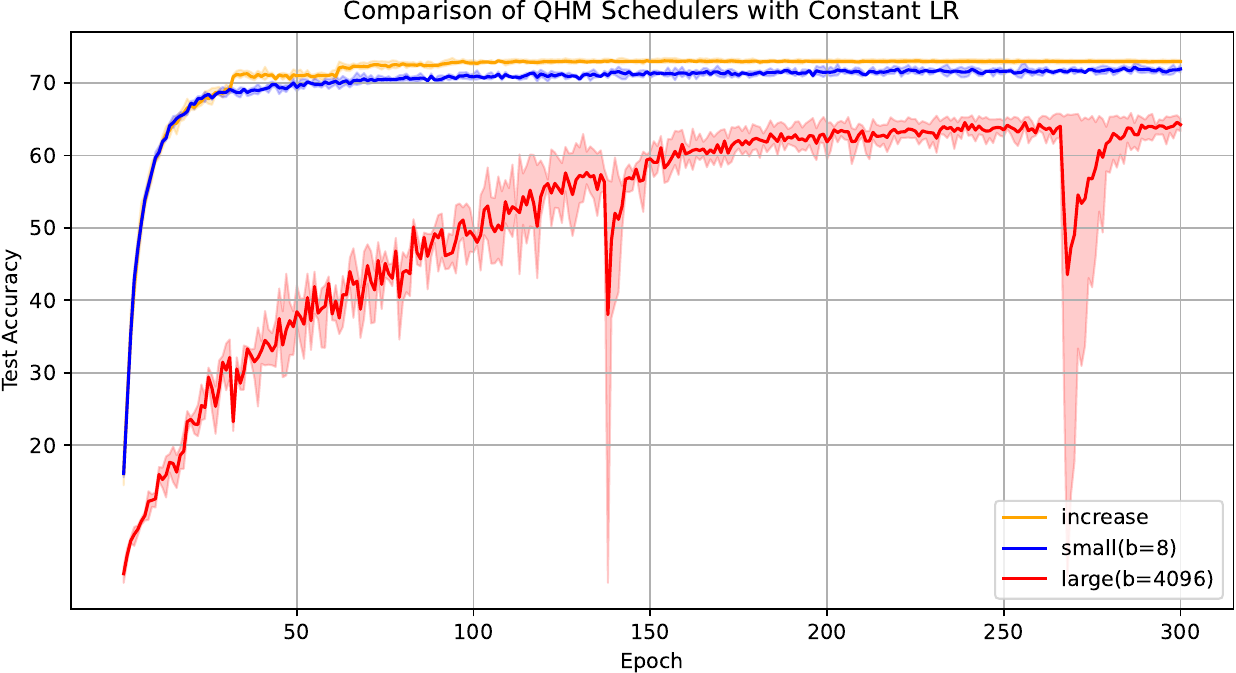}
}
\hfill
\subfigure[Decaying Squared LR]{
	\includegraphics[width=0.441\linewidth]{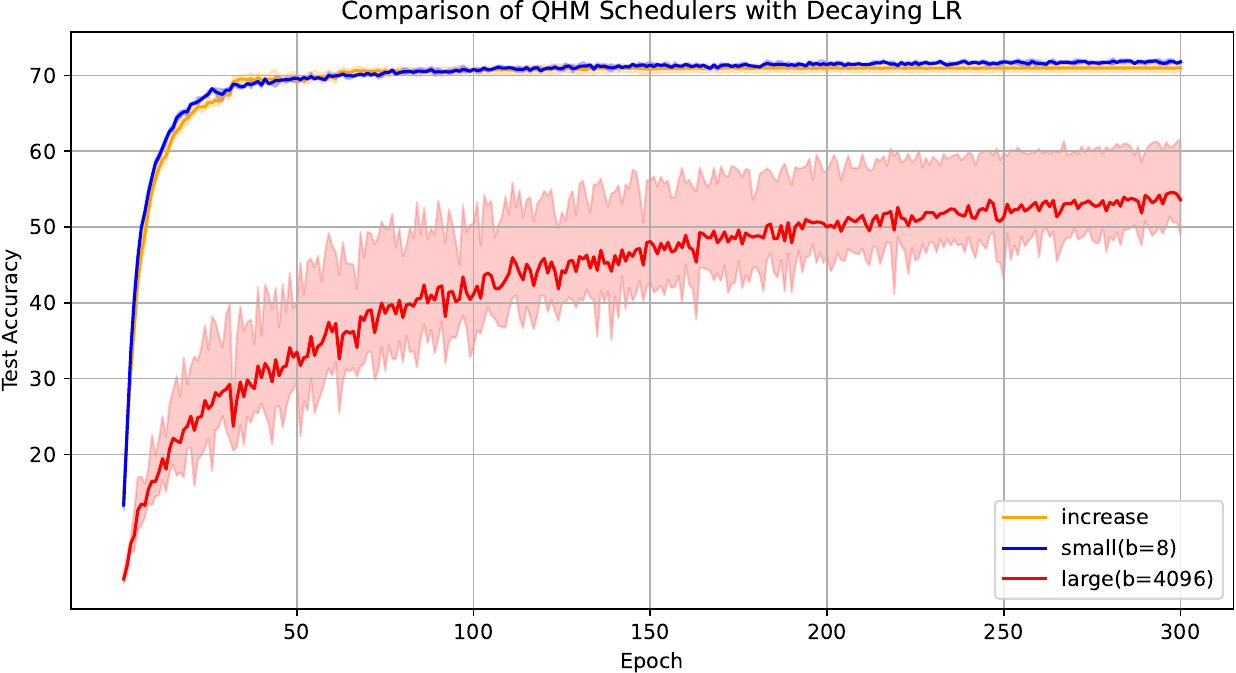}
}

\vspace{0.4cm}

\subfigure[Cosine Annealing LR]{
	\includegraphics[width=0.441\linewidth]{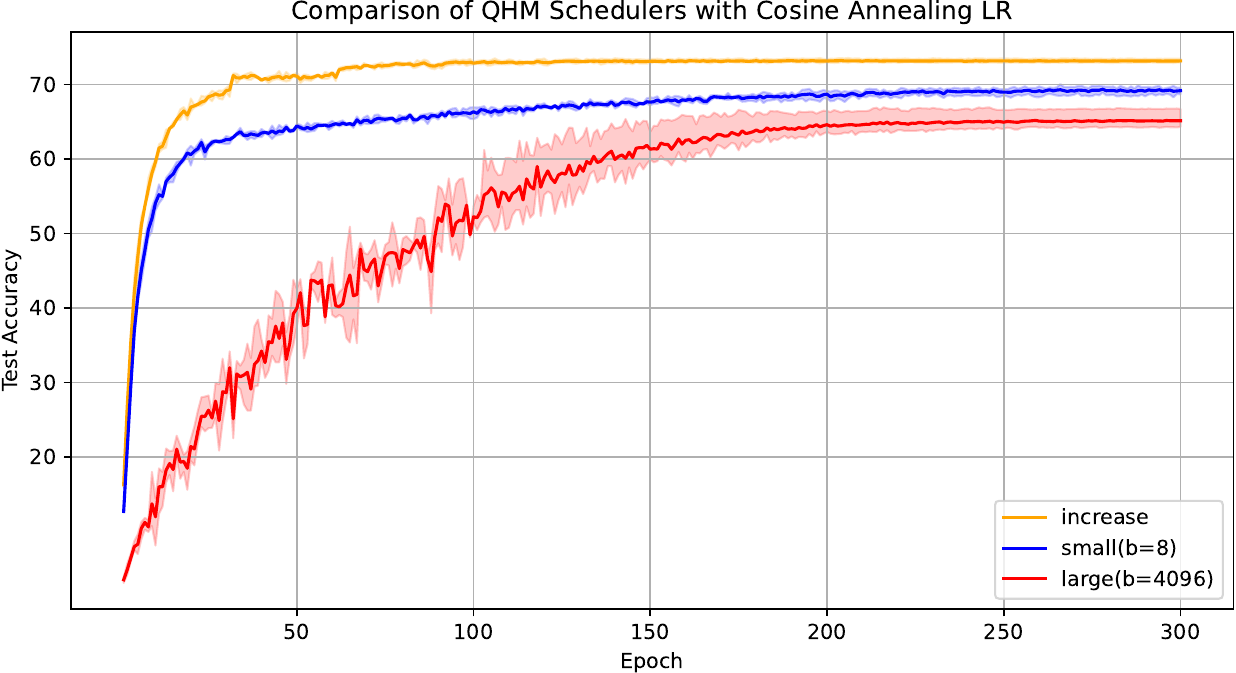}
}
\hfill
\subfigure[Polynomial Decay LR]{
	\includegraphics[width=0.441\linewidth]{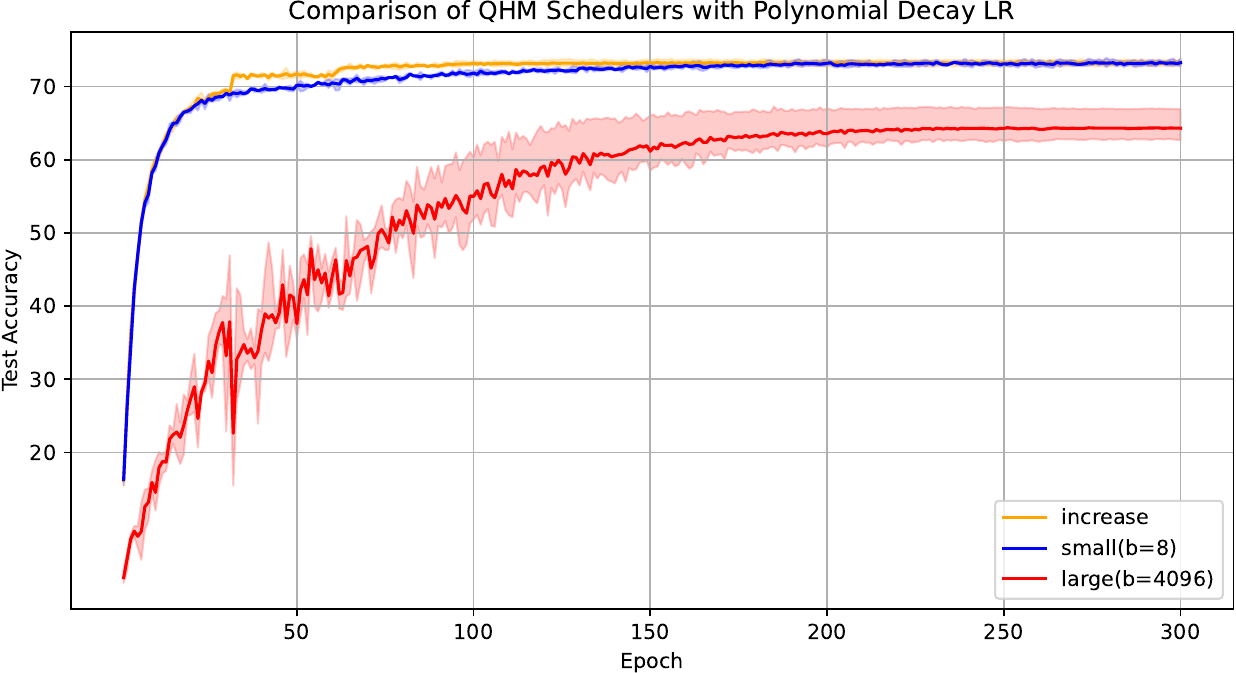}
}
\caption{Test accuracy score versus number of epochs for comparison of increasing and constant batch sizes in using mini-batch QHM with various lr-schedulers and step-decay momentum weights to train ResNet-18 on the CIFAR-100 dataset.}
\label{fig1}
\end{figure}

\begin{figure}[htbp]
\centering
\subfigure[Constant LR]{
	\includegraphics[width=0.479\linewidth]{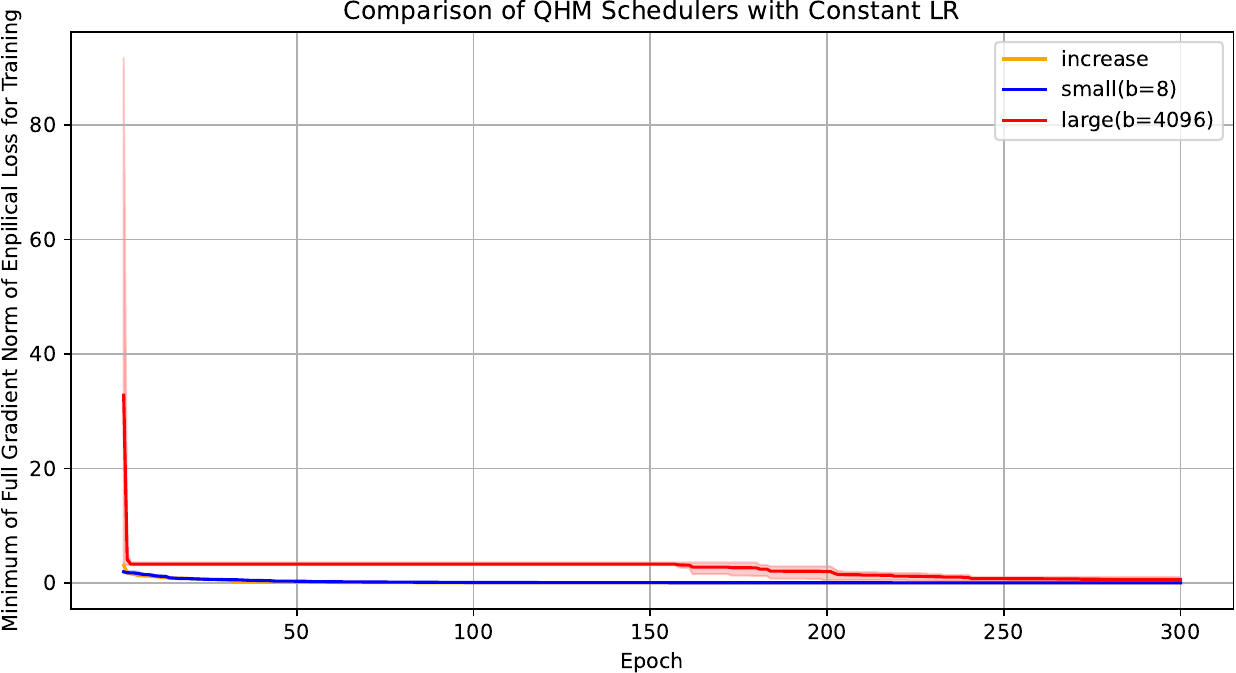}
}
\hfill
\subfigure[Decaying Squared LR]{
	\includegraphics[width=0.479\linewidth]{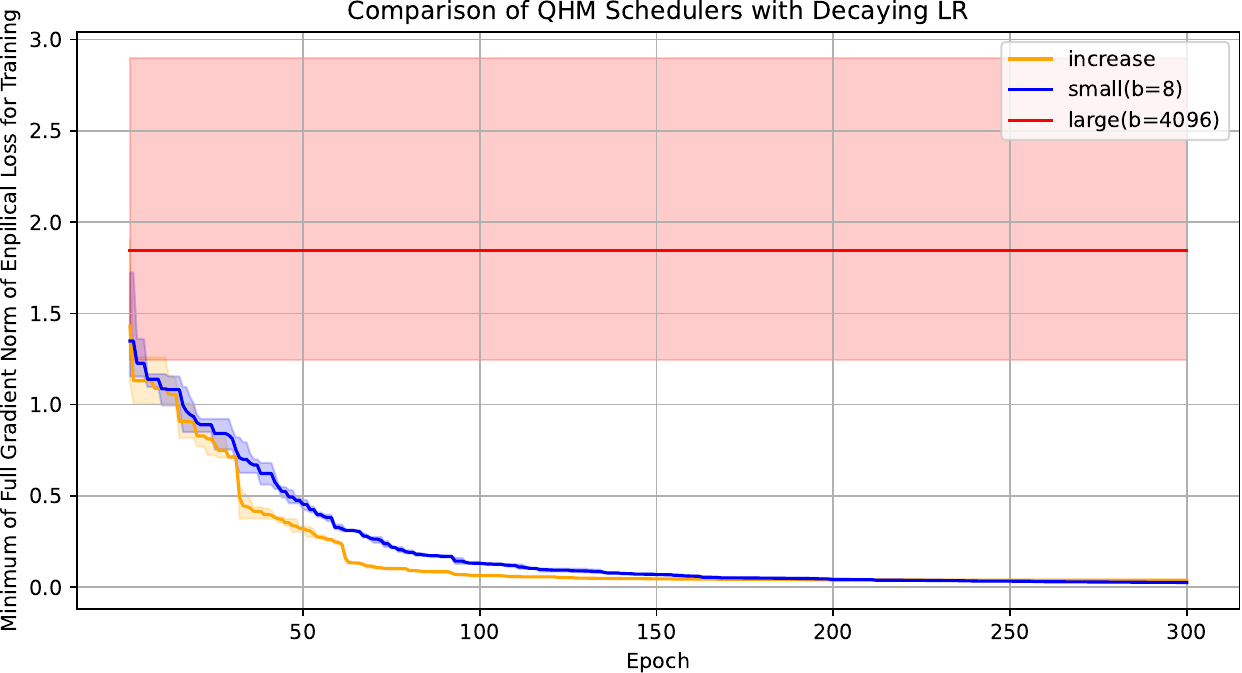}
}

\vspace{0.4cm}

\subfigure[Cosine Annealing LR]{
	\includegraphics[width=0.479\linewidth]{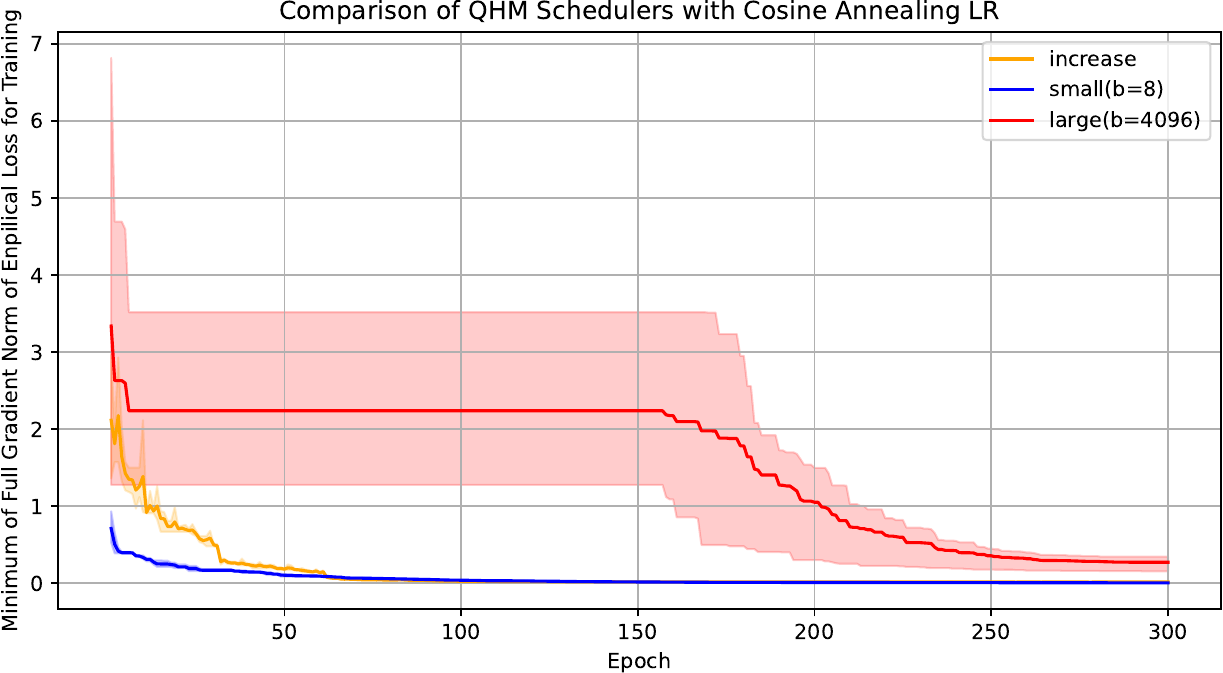}
}
\hfill
\subfigure[Polynomial Decay LR]{
	\includegraphics[width=0.479\linewidth]{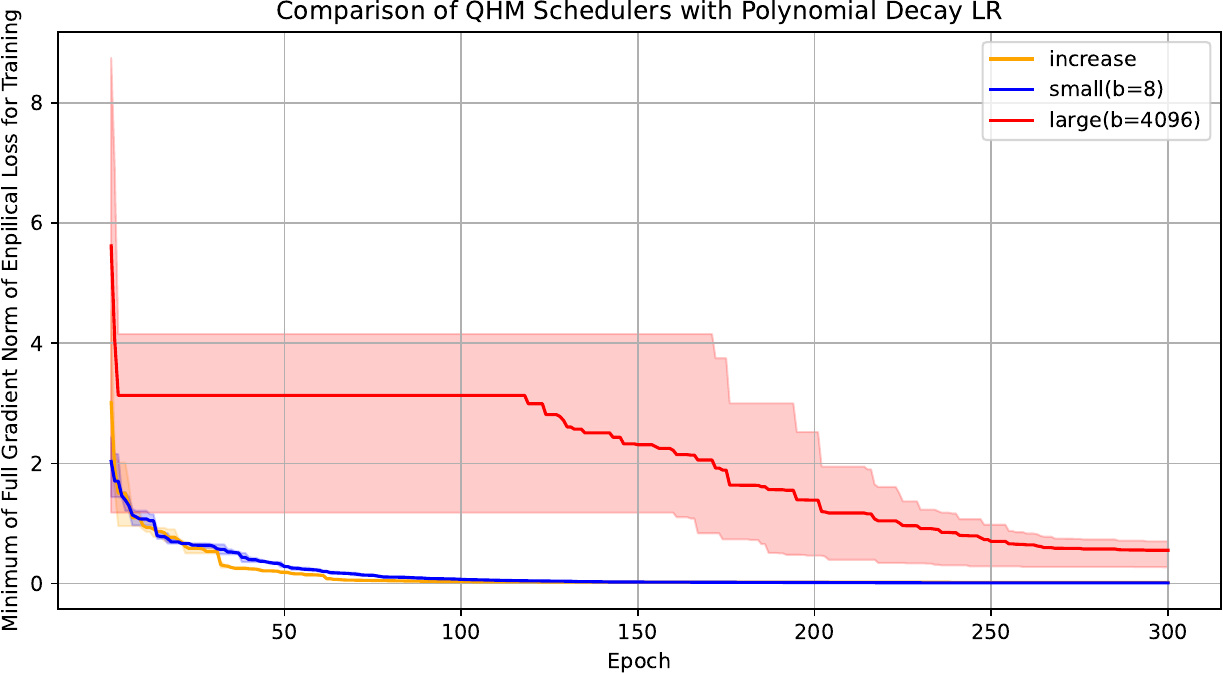}
}
\caption{Minimum of full gradient norm of empirical loss versus number of epochs for comparison of increasing and constant batch sizes in using mini-batch QHM with various lr-schedulers and step-decay momentum weights to train ResNet-18 on the CIFAR-100 dataset.}
\label{fig2}
\end{figure}

First, we verified that using increasing batch sizes leads to better performance in mini-batch QHM and NSHB in comparison with using constant batch sizes. In particular, we used a batch size doubling (i.e. $\delta = 2$ in \eqref{exp_batch}) every 30 epochs from an initial batch size $b_0 = 2^3$.
In the comparison between increasing batch sizes and constant batch sizes, one must be careful that the total number of steps $K$ differs, although the total number of $M$ is unified. In other words, as the batch size increases, the number of steps $K$ required to complete $M$ epochs decreases, which consequently leads to larger gradient norms.
Figure \ref{fig1} and \ref{fig2} indicate that the mini-batch QHM and NSHB with the increasing batch size outperformed those using the small or large constant batch sizes in terms of both test accuracy and minimum value of the full gradient norm, depending on the learning rate scheduler. In particular, it can be observed that the performance improvement is attributed to the sharp changes in test accuracy and minimum value of the full gradient norm caused by increasing the batch size at 30 and 60 epochs. In the latter stages of training, the gradual performance improvement observed with the small batch size is hypothesized to result from the larger number of steps compared to the increasing batch size. This indicates that the slightly better results of the small batch size compared to the increasing batch size (e.g., Figure \ref{fig1}(b) and Figure \ref{fig2}(a) and (c)) can be explained by the fact that the increasing batch size achieves comparable performance with fewer steps. Based on the above, these results indicate that, rather than fixing the batch size to a small or large value from the beginning, gradually increasing it from a small to a large value is beneficial for training neural networks. This finding also applies to mini-batch QHM and NSHB. Moreover, while the theoretical results required the batch size to diverge, our experiments demonstrate that even a finite increase in batch size can yield similar benefits.

\begin{figure}[htbp]
\centering
\subfigure[Minimum of full gradient norm of empirical loss versus number of epochs] {
\includegraphics[width=0.479\linewidth]{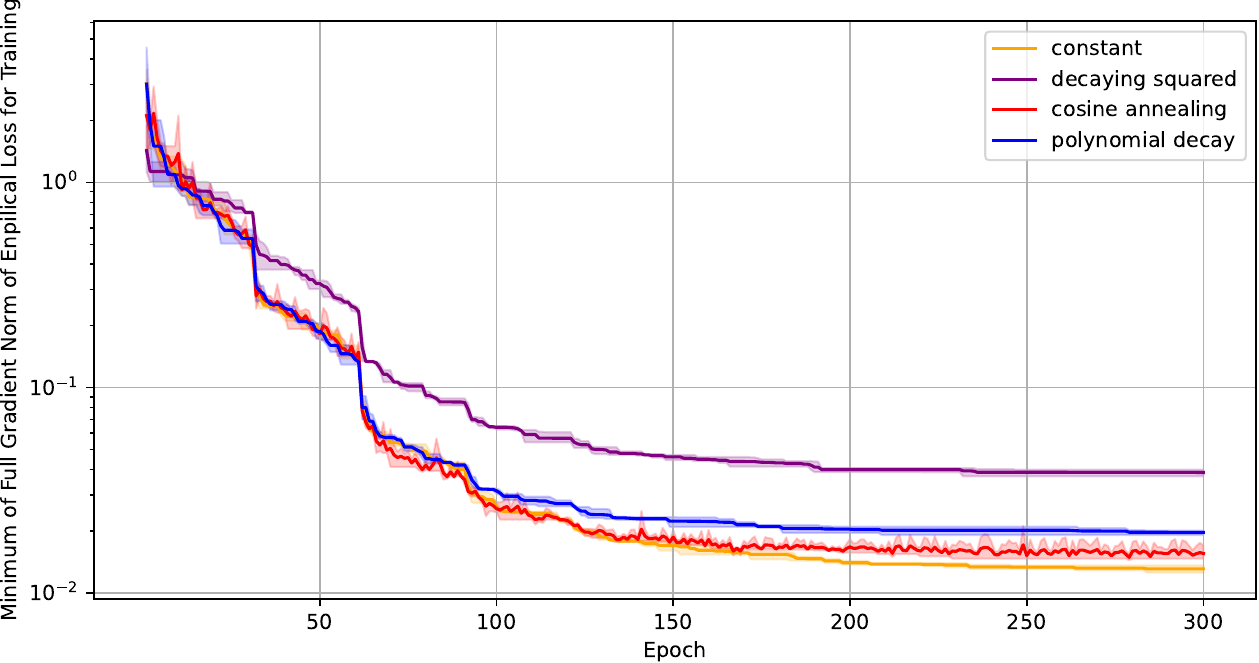}}
\hfill
\subfigure[Test accuracy score versus number of epochs] {
\includegraphics[width=0.479\linewidth]{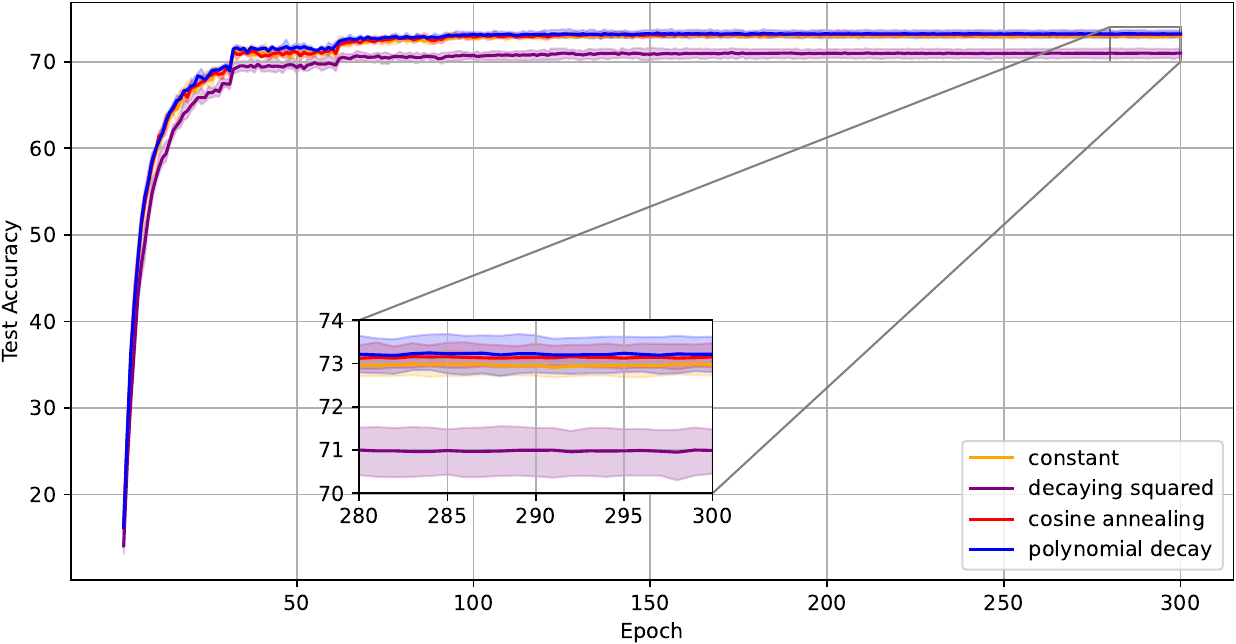}}
\caption{Comparison of lr--scheduler using mini-batch QHM with Step Decay Beta and Gamma to train ResNet-18 on the CIFAR-100 dataset.}
\label{fig3}
\end{figure}

\begin{figure}[htbp]
\centering
\subfigure[Minimum of full gradient norm of empirical loss versus number of epochs] {
\includegraphics[width=0.4\linewidth]{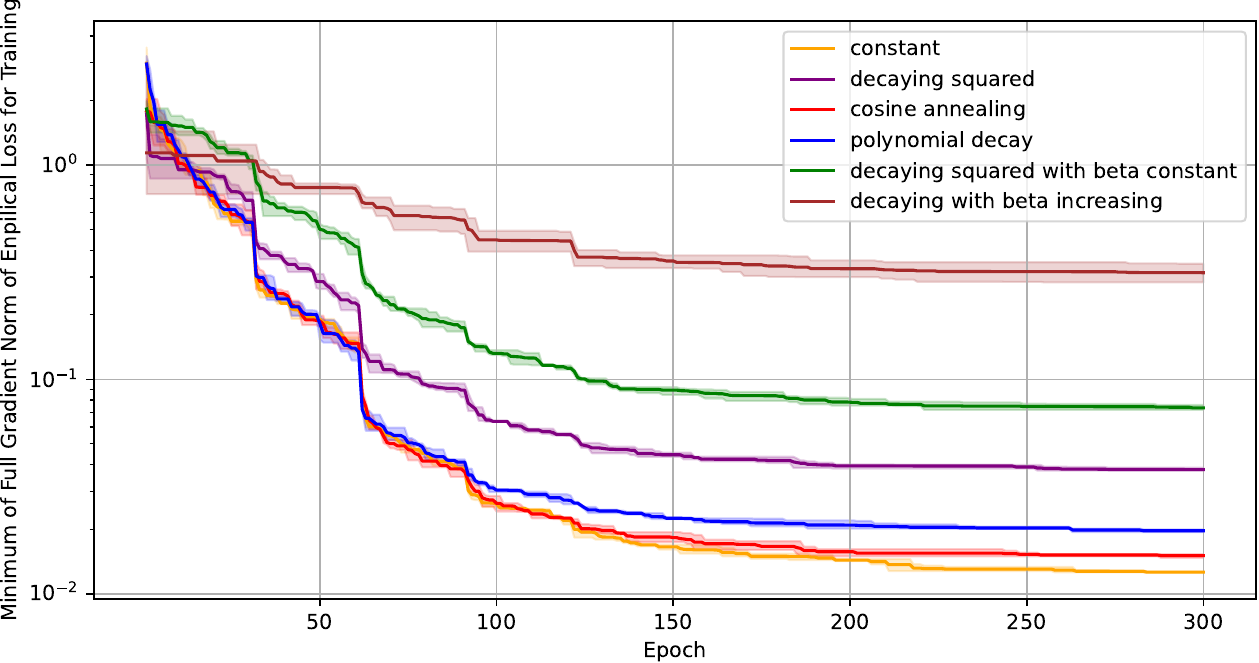}}
\hfill
\subfigure[Test accuracy score versus number of epochs] {
\includegraphics[width=0.4\linewidth]{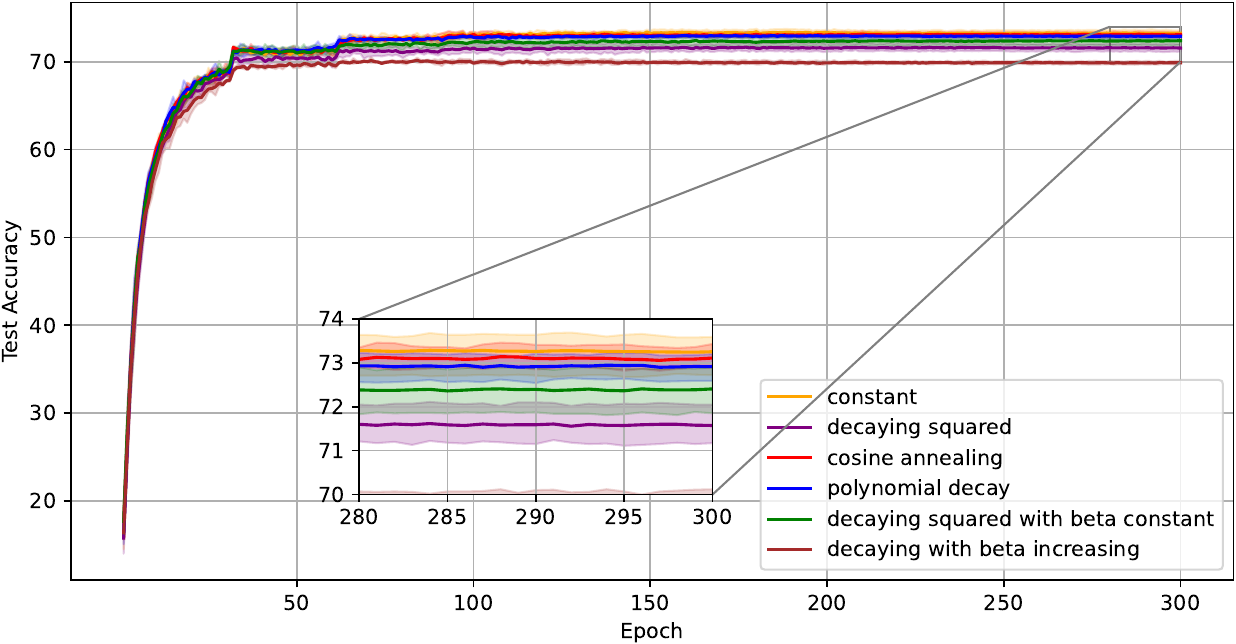}}
\caption{Comparison of lr and beta--scheduler using NSHB with Constant Gamma ($\gamma_k=1$) to train ResNet-18 on the CIFAR-100 dataset.}
\label{fig4}
\end{figure}

Next, we confirmed that [Decaying Squared LR \eqref{decaying_lr_sq}] and [Decaying LR \eqref{decaying_lr}] negatively affect training compared with the other learning-rate schedulers. Figure \ref{fig3} and \ref{fig4} compare mini-batch QHM and NSHB with increasing batch sizes under various learning rates. Both the test accuracy and the minimum gradient norm indicate that the performance of methods using [Decaying Squared LR \eqref{decaying_lr_sq}] and [Decaying LR \eqref{decaying_lr}] is worse, while [Constant LR \eqref{constant_lr}], [Cosine LR \eqref{cosine_lr}], and [Polynomial LR \eqref{poly_lr}] achieved comparable performance to what is predicted by Theorem \ref{th_1}. This holds true regardless of whether mini-batch QHM or mini-batch NSHB is used. In particular, in the case of mini-batch NSHB, although we changed the learning rate decay and the beta scheduler, the performance remained poor. Furthermore, in comparing mini-batch QHM and NSHB, we observed that their overall performances are similar. This is because, although mini-batch QHM can reduce the upper bound in Theorem \ref{th_1} more quickly, their performance in terms of non-asymptotic convergence is essentially the same.

\section{Conclusion}
This paper proved both asymptotic and non-asymptotic convergence of mini-batch QHM, NSHB, and SGD using learning rate and momentum weight schedulers with an increasing batch size. Since an increasing batch size relaxes the conditions required for asymptotic convergence, it allows us to avoid using a decaying learning rate, which is often detrimental to non-asymptotic convergence performance. While theoretically letting the batch size diverge is a very strong assumption, our numerical experiments demonstrate that even a finite increase toward this limit can still provide benefits. One limitation of this study is that the limited number of models and datasets in the experiments. Hence, we should conduct similar experiments with more models and datasets to support our theoretical results.

\bibliography{arxiv}

\begin{thebibliography}{38}
\providecommand{\natexlab}[1]{#1}
\providecommand{\url}[1]{\texttt{#1}}
\expandafter\ifx\csname urlstyle\endcsname\relax
  \providecommand{\doi}[1]{doi: #1}\else
  \providecommand{\doi}{doi: \begingroup \urlstyle{rm}\Url}\fi

\bibitem[Beck(2017)]{doi:10.1137/1.9781611974997}
Amir Beck.
\newblock \emph{First-Order Methods in Optimization}.
\newblock Society for Industrial and Applied Mathematics, Philadelphia, PA,
  2017.
\newblock \doi{10.1137/1.9781611974997}.
\newblock URL \url{https://epubs.siam.org/doi/abs/10.1137/1.9781611974997}.

\bibitem[Bollapragada et~al.(2024)Bollapragada, Chen, and
  Ward]{10.1093/imanum/drae033}
Raghu Bollapragada, Tyler Chen, and Rachel Ward.
\newblock On the fast convergence of minibatch heavy ball momentum.
\newblock \emph{IMA Journal of Numerical Analysis}, page drae033, 2024.
\newblock ISSN 0272-4979.
\newblock \doi{10.1093/imanum/drae033}.
\newblock URL \url{https://doi.org/10.1093/imanum/drae033}.

\bibitem[Dai et~al.(2016)Dai, Li, He, and Sun]{NIPS2016_577ef115}
Jifeng Dai, Yi~Li, Kaiming He, and Jian Sun.
\newblock R-fcn: Object detection via region-based fully convolutional
  networks.
\newblock In \emph{Advances in Neural Information Processing Systems},
  volume~29. Curran Associates, Inc., 2016.
\newblock URL
  \url{https://proceedings.neurips.cc/paper_files/paper/2016/file/577ef1154f3240ad5b9b413aa7346a1e-Paper.pdf}.

\bibitem[De et~al.(2017)De, Yadav, Jacobs, and Goldstein]{pmlr-v54-de17a}
Soham De, Abhay Yadav, David Jacobs, and Tom Goldstein.
\newblock {Automated Inference with Adaptive Batches}.
\newblock In \emph{Proceedings of the 20th International Conference on
  Artificial Intelligence and Statistics}, volume~54 of \emph{Proceedings of
  Machine Learning Research}, pages 1504--1513. PMLR, 2017.
\newblock URL \url{https://proceedings.mlr.press/v54/de17a.html}.

\bibitem[Ghadimi et~al.(2016)Ghadimi, Lan, and Zhang]{Ghadimi:2016aa}
Saeed Ghadimi, Guanghui Lan, and Hongchao Zhang.
\newblock Mini-batch stochastic approximation methods for nonconvex stochastic
  composite optimization.
\newblock \emph{Mathematical Programming}, 155\penalty0 (1):\penalty0 267--305,
  2016.
\newblock \doi{10.1007/s10107-014-0846-1}.
\newblock URL \url{https://doi.org/10.1007/s10107-014-0846-1}.

\bibitem[Gitman et~al.(2019)Gitman, Lang, Zhang, and
  Xiao]{NEURIPS2019_4eff0720}
Igor Gitman, Hunter Lang, Pengchuan Zhang, and Lin Xiao.
\newblock Understanding the role of momentum in stochastic gradient methods.
\newblock In \emph{Advances in Neural Information Processing Systems},
  volume~32. Curran Associates, Inc., 2019.
\newblock URL
  \url{https://proceedings.neurips.cc/paper_files/paper/2019/file/4eff0720836a198b6174eecf02cbfdbf-Paper.pdf}.

\bibitem[Gupal and Bazhenov(1972)]{nshb}
A.~Gupal and L.~T. Bazhenov.
\newblock A stochastic analog of the conjugate gradient method.
\newblock \emph{Cybernetics}, 8\penalty0 (1):\penalty0 138--140, 1972.

\bibitem[Habibian et~al.(2021)Habibian, Abati, Cohen, and Bejnordi]{skipconv}
Amirhossein Habibian, Davide Abati, Taco Cohen, and Babak~Ehteshami Bejnordi.
\newblock Skip-convolutions for efficient video processing.
\newblock In \emph{Proceedings of the IEEE Conference on Computer Vision and
  Pattern Recognition}, 2021.

\bibitem[He et~al.(2015)He, Zhang, Ren, and Sun]{He2015DeepRL}
Kaiming He, X.~Zhang, Shaoqing Ren, and Jian Sun.
\newblock Deep residual learning for image recognition.
\newblock \emph{2016 IEEE Conference on Computer Vision and Pattern Recognition
  (CVPR)}, pages 770--778, 2015.
\newblock URL \url{https://api.semanticscholar.org/CorpusID:206594692}.

\bibitem[Hinton et~al.(2012)Hinton, Deng, Yu, Dahl, Mohamed, Jaitly, Senior,
  Vanhoucke, Nguyen, Sainath, and Kingsbury]{6296526}
Geoffrey Hinton, Li~Deng, Dong Yu, George~E. Dahl, Abdel-rahman Mohamed,
  Navdeep Jaitly, Andrew Senior, Vincent Vanhoucke, Patrick Nguyen, Tara~N.
  Sainath, and Brian Kingsbury.
\newblock Deep neural networks for acoustic modeling in speech recognition: The
  shared views of four research groups.
\newblock \emph{IEEE Signal Processing Magazine}, 29\penalty0 (6):\penalty0
  82--97, 2012.
\newblock \doi{10.1109/MSP.2012.2205597}.

\bibitem[Hoffer et~al.(2017)Hoffer, Hubara, and Soudry]{NIPS2017_a5e0ff62}
Elad Hoffer, Itay Hubara, and Daniel Soudry.
\newblock Train longer, generalize better: closing the generalization gap in
  large batch training of neural networks.
\newblock In \emph{Advances in Neural Information Processing Systems},
  volume~30, 2017.
\newblock URL
  \url{https://proceedings.neurips.cc/paper_files/paper/2017/file/a5e0ff62be0b08456fc7f1e88812af3d-Paper.pdf}.

\bibitem[Imaizumi and Iiduka(2024)]{Imaizumi19062024}
Kento Imaizumi and Hideaki Iiduka.
\newblock Iteration and stochastic first-order oracle complexities of
  stochastic gradient descent using constant and decaying learning rates.
\newblock \emph{Optimization}, 0\penalty0 (0):\penalty0 1--24, 2024.
\newblock \doi{10.1080/02331934.2024.2367635}.

\bibitem[Keskar et~al.(2017)Keskar, Mudigere, Nocedal, Smelyanskiy, and
  Tang]{keskar2017on}
Nitish~Shirish Keskar, Dheevatsa Mudigere, Jorge Nocedal, Mikhail Smelyanskiy,
  and Ping Tak~Peter Tang.
\newblock On large-batch training for deep learning: Generalization gap and
  sharp minima.
\newblock In \emph{International Conference on Learning Representations}, 2017.
\newblock URL \url{https://openreview.net/forum?id=H1oyRlYgg}.

\bibitem[Kingma and Ba(2015)]{adam}
Diederik~P. Kingma and Jimmy Ba.
\newblock Adam: A method for stochastic optimization.
\newblock In \emph{Proceedings of The International Conference on Learning
  Representations}, 2015.

\bibitem[Krizhevsky(2009)]{Krizhevsky2009LearningML}
Alex Krizhevsky.
\newblock Learning multiple layers of features from tiny images.
\newblock 2009.
\newblock URL \url{https://api.semanticscholar.org/CorpusID:18268744}.

\bibitem[Krizhevsky et~al.(2012)Krizhevsky, Sutskever, and
  Hinton]{krizhevsky2012imagenet}
Alex Krizhevsky, Ilya Sutskever, and Geoffrey~E Hinton.
\newblock Imagenet classification with deep convolutional neural networks.
\newblock In \emph{Advances in Neural Information Processing Systems},
  volume~25, pages 1097--1105. Curran Associates, Inc., 2012.
\newblock URL
  \url{https://proceedings.neurips.cc/paper/2012/file/c399862d3b9d6b76c8436e924a68c45b-Paper.pdf}.

\bibitem[Li et~al.(2024)Li, Zhao, Zhang, Sun, Wu, Jiao, Wang, Liu, Fang, Xue,
  Tao, Cui, and Wang]{NEURIPS2024_ef74413c}
Shuaipeng Li, Penghao Zhao, Hailin Zhang, Xingwu Sun, Hao Wu, Dian Jiao, Weiyan
  Wang, Chengjun Liu, Zheng Fang, Jinbao Xue, Yangyu Tao, Bin Cui, and Di~Wang.
\newblock Surge phenomenon in optimal learning rate and batch size scaling.
\newblock In \emph{Advances in Neural Information Processing Systems},
  volume~37, pages 132722--132746. Curran Associates, Inc., 2024.
\newblock URL
  \url{https://proceedings.neurips.cc/paper_files/paper/2024/file/ef74413c7bf1d915c3e45c72e19a5d32-Paper-Conference.pdf}.

\bibitem[Liu et~al.(2020)Liu, Gao, and Yin]{NEURIPS2020_d3f5d4de}
Yanli Liu, Yuan Gao, and Wotao Yin.
\newblock An improved analysis of stochastic gradient descent with momentum.
\newblock In \emph{Advances in Neural Information Processing Systems},
  volume~33, pages 18261--18271. Curran Associates, Inc., 2020.

\bibitem[Loshchilov and Hutter(2017)]{loshchilov2017sgdr}
Ilya Loshchilov and Frank Hutter.
\newblock {SGDR}: Stochastic gradient descent with warm restarts.
\newblock In \emph{International Conference on Learning Representations}, 2017.

\bibitem[Loshchilov and Hutter(2019)]{loshchilov2018decoupled}
Ilya Loshchilov and Frank Hutter.
\newblock Decoupled weight decay regularization.
\newblock In \emph{Proceedings of The International Conference on Learning
  Representations}, 2019.

\bibitem[Ma and Yarats(2019)]{ma2018quasihyperbolic}
Jerry Ma and Denis Yarats.
\newblock Quasi-hyperbolic momentum and adam for deep learning.
\newblock In \emph{International Conference on Learning Representations}, 2019.

\bibitem[Nesterov(1983)]{Nesterov1983AMF}
Yurii Nesterov.
\newblock A method for solving the convex programming problem with convergence
  rate o($1/k^2$).
\newblock \emph{Proceedings of the USSR Academy of Sciences}, 269:\penalty0
  543--547, 1983.
\newblock URL \url{https://api.semanticscholar.org/CorpusID:145918791}.

\bibitem[Polyak(1964)]{polyak1964}
Boris~T. Polyak.
\newblock Some methods of speeding up the convergence of iteration methods.
\newblock \emph{USSR Computational Mathematics and Mathematical Physics},
  4:\penalty0 1--17, 1964.

\bibitem[Redmon et~al.(2016)Redmon, Divvala, Girshick, and Farhadi]{7780460}
Joseph Redmon, Santosh Divvala, Ross Girshick, and Ali Farhadi.
\newblock You only look once: Unified, real-time object detection.
\newblock In \emph{2016 IEEE Conference on Computer Vision and Pattern
  Recognition (CVPR)}, pages 779--788, 2016.
\newblock \doi{10.1109/CVPR.2016.91}.

\bibitem[Robbins and Monro(1951)]{10.1214/aoms/1177729586}
Herbert Robbins and Sutton Monro.
\newblock {A Stochastic Approximation Method}.
\newblock \emph{The Annals of Mathematical Statistics}, 22\penalty0
  (3):\penalty0 400 -- 407, 1951.
\newblock \doi{10.1214/aoms/1177729586}.
\newblock URL \url{https://doi.org/10.1214/aoms/1177729586}.

\bibitem[Sato and Iiduka(2023)]{pmlr-v202-sato23b}
Naoki Sato and Hideaki Iiduka.
\newblock Existence and estimation of critical batch size for training
  generative adversarial networks with two time-scale update rule.
\newblock In \emph{Proceedings of the 40th International Conference on Machine
  Learning}, volume 202 of \emph{Proceedings of Machine Learning Research},
  pages 30080--30104. PMLR, 2023.
\newblock URL \url{https://proceedings.mlr.press/v202/sato23b.html}.

\bibitem[Sato and Iiduka(2025)]{Sato_Iiduka_2025}
Naoki Sato and Hideaki Iiduka.
\newblock Explicit and implicit graduated optimization in deep neural networks.
\newblock \emph{Proceedings of the AAAI Conference on Artificial Intelligence},
  39\penalty0 (19):\penalty0 20283--20291, 2025.
\newblock \doi{10.1609/aaai.v39i19.34234}.
\newblock URL \url{https://ojs.aaai.org/index.php/AAAI/article/view/34234}.

\bibitem[Shallue et~al.(2019)Shallue, Lee, Antognini, Sohl-Dickstein, Frostig,
  and Dahl]{shallue2019}
Christopher~J. Shallue, Jaehoon Lee, Joseph Antognini, Jascha Sohl-Dickstein,
  Roy Frostig, and George~E. Dahl.
\newblock Measuring the effects of data parallelism on neural network training.
\newblock \emph{Journal of Machine Learning Research}, 20:\penalty0 1--49,
  2019.

\bibitem[Smith et~al.(2018)Smith, Kindermans, and Le]{l.2018dont}
Samuel~L. Smith, Pieter-Jan Kindermans, and Quoc~V. Le.
\newblock Don't decay the learning rate, increase the batch size.
\newblock In \emph{International Conference on Learning Representations}, 2018.

\bibitem[Sun et~al.(2021)Sun, Yang, Xun, and Zhang]{10.1145/3447548.3467287}
Jianhui Sun, Ying Yang, Guangxu Xun, and Aidong Zhang.
\newblock A stagewise hyperparameter scheduler to improve generalization.
\newblock In \emph{Proceedings of the 27th ACM SIGKDD Conference on Knowledge
  Discovery \& Data Mining}, KDD '21, page 1530–1540, New York, NY, USA,
  2021. Association for Computing Machinery.
\newblock ISBN 9781450383325.
\newblock \doi{10.1145/3447548.3467287}.
\newblock URL \url{https://doi.org/10.1145/3447548.3467287}.

\bibitem[Sutskever et~al.(2013)Sutskever, Martens, Dahl, and
  Hinton]{pmlr-v28-sutskever13}
Ilya Sutskever, James Martens, George Dahl, and Geoffrey Hinton.
\newblock On the importance of initialization and momentum in deep learning.
\newblock In \emph{Proceedings of the 30th International Conference on Machine
  Learning}, volume~28 of \emph{Proceedings of Machine Learning Research},
  pages 1139--1147, Atlanta, Georgia, USA, 2013. PMLR.
\newblock URL \url{https://proceedings.mlr.press/v28/sutskever13.html}.

\bibitem[Tieleman and Hinton(2012)]{rmsprop}
Tijmen Tieleman and Geoffrey Hinton.
\newblock {RMSP}rop: {D}ivide the gradient by a running average of its recent
  magnitude.
\newblock \emph{{COURSERA}: {N}eural networks for machine learning},
  4:\penalty0 26--31, 2012.

\bibitem[Umeda and Iiduka(2025)]{umeda2025increasing}
Hikaru Umeda and Hideaki Iiduka.
\newblock Increasing both batch size and learning rate accelerates stochastic
  gradient descent.
\newblock \emph{Transactions on Machine Learning Research}, 2025.
\newblock ISSN 2835-8856.

\bibitem[Yan et~al.(2018)Yan, Yang, Li, Lin, and Yang]{ijcai2018p410}
Yan Yan, Tianbao Yang, Zhe Li, Qihang Lin, and Yi~Yang.
\newblock A unified analysis of stochastic momentum methods for deep learning.
\newblock In \emph{Proceedings of the Twenty-Seventh International Joint
  Conference on Artificial Intelligence, {IJCAI-18}}, pages 2955--2961.
  International Joint Conferences on Artificial Intelligence Organization,
  2018.
\newblock \doi{10.24963/ijcai.2018/410}.
\newblock URL \url{https://doi.org/10.24963/ijcai.2018/410}.

\bibitem[Yang et~al.(2016)Yang, Lin, and
  Li]{yang2016unifiedconvergenceanalysisstochastic}
Tianbao Yang, Qihang Lin, and Zhe Li.
\newblock Unified convergence analysis of stochastic momentum methods for
  convex and non-convex optimization, 2016.
\newblock URL \url{https://arxiv.org/abs/1604.03257}.

\bibitem[You et~al.(2017)You, Gitman, and
  Ginsburg]{you2017largebatchtrainingconvolutional}
Yang You, Igor Gitman, and Boris Ginsburg.
\newblock Large batch training of convolutional networks, 2017.
\newblock URL \url{https://arxiv.org/abs/1708.03888}.

\bibitem[You et~al.(2020)You, Li, Reddi, Hseu, Kumar, Bhojanapalli, Song,
  Demmel, Keutzer, and Hsieh]{You2020Large}
Yang You, Jing Li, Sashank Reddi, Jonathan Hseu, Sanjiv Kumar, Srinadh
  Bhojanapalli, Xiaodan Song, James Demmel, Kurt Keutzer, and Cho-Jui Hsieh.
\newblock Large batch optimization for deep learning: Training bert in 76
  minutes.
\newblock In \emph{International Conference on Learning Representations}, 2020.
\newblock URL \url{https://openreview.net/forum?id=Syx4wnEtvH}.

\bibitem[Yu et~al.(2019)Yu, Jin, and Yang]{pmlr-v97-yu19d}
Hao Yu, Rong Jin, and Sen Yang.
\newblock On the linear speedup analysis of communication efficient momentum
  {SGD} for distributed non-convex optimization.
\newblock In \emph{Proceedings of the 36th International Conference on Machine
  Learning}, volume~97 of \emph{Proceedings of Machine Learning Research},
  pages 7184--7193. PMLR, 2019.
\newblock URL \url{https://proceedings.mlr.press/v97/yu19d.html}.

\end{thebibliography}

\newpage
\appendix

\section{Appendix}
\subsection{Definition of Normalized Stochastic Heavy Ball and Stochastic Heavy Ball}

\begin{algorithm} 
\caption{Stochastic Heavy Ball (SHB)} 
\label{algo_shb} 
\begin{algorithmic}[1] 
\REQUIRE
$\bm{x}_{0} \in\mathbb{R}^d$ (initial point), $\hat{\alpha}_k \in (0,+\infty)$ (learning rate), $\hat{\beta}_k \in [0,+\infty)$ (momentum parameter), $b_k \in \mathbb{N}$ (batch size), $K \geq 1$ (steps)
\ENSURE
$\bm{x}_{K}$
\FOR{$k = 0, 1, \ldots, K-1$}
\STATE
$\nabla f_{B_k} (\bm{x}_k)
:= \frac{1}{b_k} \sum_{i\in [b_k]} \nabla f_{\xi_{k,i}}(\bm{x}_k)$
\STATE
$\bm{m}_k := \nabla f_{B_k}(\bm{x}_k) + \hat{\beta}_k\bm{m}_{k-1}$
\STATE
$\bm{x}_{k+1} := \bm{x}_k - \hat{\alpha}_k\bm{m}_k$
\ENDFOR 
\end{algorithmic}
\end{algorithm}

\begin{algorithm} 
\caption{Normalized Stochastic Heavy Ball (NSHB)} 
\label{algo_nshb} 
\begin{algorithmic}[1] 
\REQUIRE
$\bm{x}_{0} \in\mathbb{R}^d$ (initial point), $\tilde{\alpha}_k \in (0,+\infty)$ (learning rate), $\tilde{\beta}_k \in [0,1]$ (momentum parameter), $b_k \in \mathbb{N}$ (batch size), $K \geq 1$ (steps)
\ENSURE
$\bm{x}_{K}$
\FOR{$k = 0, 1, \ldots, K-1$}
\STATE
$\nabla f_{B_k} (\bm{x}_k)
:= \frac{1}{b_k} \sum_{i\in [b_k]} \nabla f_{\xi_{k,i}}(\bm{x}_k)$
\STATE
$\bm{m}_k := (1 - \tilde{\beta}_k)\nabla f_{B_k}(\bm{x}_k) + \tilde{\beta}_k\bm{m}_{k-1}$
\STATE
$\bm{x}_{k+1} := \bm{x}_k - \tilde{\alpha}_k\bm{m}_k$
\ENDFOR 
\end{algorithmic}
\end{algorithm}

\subsection{Relationship between SHB and NSHB} 
\label{shb_and_nshb}
\begin{proposition}
(i) When we use SHB, SHB becomes NSHB by setting
\begin{align*}
\hat{\alpha}_k = \tilde{\alpha}_k(1 - \tilde{\beta}_k), \quad \hat{\beta}_k = \frac{\tilde{\beta}_k}{1-\tilde{\beta}_k}.
\end{align*}
(ii) When we use NSHB, NSHB becomes SHB by setting
\begin{align*}
\tilde{\alpha}_k = \hat{\alpha}_k(1 + \hat{\beta}_k), \quad \tilde{\beta}_k = \frac{\hat{\beta}_k}{1+\hat{\beta}_k}.
\end{align*}
\end{proposition}

\proof
From the update rule of SHB and NSHB, they are also expressed as follows:
\begin{align*}
&\text{SHB: } \bm{x}_{k+1} := \bm{x}_k - \hat{\alpha}_k\nabla f_{B_k}(\bm{x}_k) - \hat{\alpha}_k\hat{\beta}_k\bm{m}_{k-1}, \\
&\text{NSHB: } \bm{x}_{k+1} := \bm{x}_k - \tilde{\alpha}_k(1 - \tilde{\beta}_k)\nabla f_{B_k}(\bm{x}_k) - \tilde{\alpha}_k\tilde{\beta}_k\bm{m}_{k-1}.
\end{align*}

Therefore, the fact that SHB is equal to NSHB, it is sufficient that $\hat{\alpha}_k = \tilde{\alpha}_k(1 - \tilde{\beta}_k)$ and $\hat{\alpha}_k\hat{\beta}_k = \tilde{\alpha}_k\tilde{\beta}_k$ hold. These equations can be solved separately for $\hat{\alpha}_k$ and $\hat{\beta}_k$, and for $\tilde{\alpha}_k$ and $\tilde{\beta}_k$. These proofs have thus been completed.
\endproof

\subsection{Lemma}
First, we will prove the following lemma.

\begin{lemma} \label{bound}
for all $k \in \mathbb{N}_0$,
\begin{align*}
\| \nabla f_{B_k}(\bm{x}_k) \| \leq G, \quad \| \bm{d}_{k} \| \leq G, \quad \quad \| \bm{m}_{k} \| \leq G.
\end{align*}
\end{lemma}

\proof

First, we will prove that, for all $k \in \mathbb{N}_0$, 
\begin{align} \label{b_f_B}
\| \nabla f_{B_k}(\bm{x}_k) \| \leq G.
\end{align}

From Condition (C3) and the triangle inequality, we find that, for all $k \in \mathbb{N}_0$,

\begin{align} \label{f_B}
\| \nabla f_{B_k}(\bm{x}_k) \| \leq \frac{1}{b_k}\sum_{i=1}^{b_k}\| \nabla f_{\xi_{k, i}}(\bm{x}_k)\|.
\end{align}

Since $\xi_{k, i}$ is a random variable sampled from the index set $[n]$, we find that, for all $k \in \mathbb{N}_0$ and $i \in [n]$, there exists $j \in [n]$ such that

\begin{align} \label{b_G}
\| \nabla f_{\xi_{k, i}}(\bm{x}_k)\| = \| \nabla f_j(\bm{x}_k) \| \leq G.
\end{align}

Substituting \eqref{b_G} for \eqref{f_B} implies that \eqref{b_f_B} holds.

Second, we will prove that, for all $k \in \mathbb{N}_0$,
\begin{align} \label{b_d}
\| \bm{d}_{k} \| \leq G.
\end{align}

In the case of $k = 0$, from the update rule of mini-batch QHM and the triangle inequality, we have 

\begin{align}
\| \bm{d}_0 \| \leq (1 - \beta_0)\| \nabla f_{B_0}(\bm{x}_0) \| + \beta_0\| \bm{d}_{-1} \| \leq (1 - \beta_0)G \leq G,
\end{align}

where $\bm{d}_{-1} = \bm{0}$. In the case of $k \in \mathbb{N}$, we assume \eqref{b_d}. Then, as above, we obtain 
\begin{align*}
\| \bm{d}_{k+1} \| &\leq (1-\beta_k)\|\nabla f_{B_k}(\bm{x}_k)\| + \beta_k\| \bm{d}_k \| \\
&\leq (1-\beta_k)G + \beta_kG = G.
\end{align*}

Therefore, \eqref{b_d} holds from mathematical induction. $\| \bm{m}_{k} \| \leq G$ can also be proved by mathematical induction. These proofs have thus been completed.

\endproof

Next, we will prove the following lemma using the proof of Theorem \ref{th_2}.

\begin{lemma} \label{lem_2}
The sequence $\{\bm{x}_k\}_{k \in \mathbb{N}_0}$ is generated by Algorithm \ref{algo_qhm} under (C1) -- (C3) satisfies
\begin{align*}
\mathbb{E}\left[\left\| \bm{d}_k - \nabla f(\bm{x}_{k+1}) \right\|^2 | \bm{x}_k \right]
&\leq \beta_k\left\| \bm{d}_{k-1} - \nabla f(\bm{x}_k)\right\|^2 + 2(1-\beta_k)^2\frac{\sigma^2}{b_k} + \alpha_k^2L^2\left(2 + \frac{1}{1 -\beta_k}\right)\mathbb{E}\left[\|\bm{m}_k\|^2 | \bm{x}_k \right],
\end{align*}
where $\mathbb{E}\left[ \ \cdot \ | \ \bm{x}_k\right]$ is the expectation conditioned on $\bm{x}_k$.
\end{lemma}

\proof

The update rule of mini-batch QHM ensures that, for all $k \in \mathbb{N}_0$,
\begin{align*}
\bm{d}_k - \nabla f(\bm{x}_{k+1}) &= (1 - \beta_k)\nabla f_{B_k}(\bm{x}_k) + \beta_k\bm{d}_{k-1} - \nabla f(\bm{x}_{k+1}) + \nabla f(\bm{x}_{k}) - \nabla f(\bm{x}_{k}) \\
&= \beta_k\left(\bm{d}_{k-1} - \nabla f(\bm{x}_{k})\right) + (1 - \beta_k)\left(\nabla f_{B_k}(\bm{x}_k) - \nabla f(\bm{x}_{k})\right) + \left(\nabla f(\bm{x}_{k}) - \nabla f(\bm{x}_{k+1})\right).
\end{align*}

Therefore, for all $k \in \mathbb{N}_0$,
\begin{align}
\begin{split} \label{d}
\left\| \bm{d}_k - \nabla f(\bm{x}_{k+1}) \right\|^2 &= \beta_k^2\left\| \bm{d}_{k-1} - \nabla f(\bm{x}_k)\right\|^2 + \left\|(1 - \beta_k)\left(\nabla f_{B_k}(\bm{x}_k) - \nabla f(\bm{x}_{k})\right) + \left(\nabla f(\bm{x}_{k}) - \nabla f(\bm{x}_{k+1})\right)\right\|^2 \\
& \quad + 2\beta_k\left\langle\bm{d}_{k-1} - \nabla f(\bm{x}_k), (1 - \beta_k)\left(\nabla f_{B_k}(\bm{x}_k) - \nabla f(\bm{x}_{k})\right) + \left(\nabla f(\bm{x}_{k}) - \nabla f(\bm{x}_{k+1})\right)\right\rangle \\
&\leq \beta_k^2\left\| \bm{d}_{k-1} - \nabla f(\bm{x}_k)\right\|^2 + 2(1-\beta_k)^2\left\|\nabla f_{B_k}(\bm{x}_k) - \nabla f(\bm{x}_{k})\right\|^2 \\
& \quad + 2\left\|f(\bm{x}_{k}) - \nabla f(\bm{x}_{k+1})\right\|^2 + 2\beta_k\left\langle\bm{d}_{k-1} - \nabla f(\bm{x}_k), (1 - \beta_k)\left(\nabla f_{B_k}(\bm{x}_k) - \nabla f(\bm{x}_{k})\right)\right\rangle \\
& \quad + 2\beta_k\left\langle\bm{d}_{k-1} - \nabla f(\bm{x}_k), f(\bm{x}_{k}) - \nabla f(\bm{x}_{k+1})\right\rangle \\
&\leq \beta_k^2\left\| \bm{d}_{k-1} - \nabla f(\bm{x}_k)\right\|^2 + 2(1-\beta_k)^2\left\|\nabla f_{B_k}(\bm{x}_k) - \nabla f(\bm{x}_{k})\right\|^2 \\
& \quad + 2\left\|f(\bm{x}_{k}) - \nabla f(\bm{x}_{k+1})\right\|^2 + 2\beta_k\left\langle\bm{d}_{k-1} - \nabla f(\bm{x}_k), (1 - \beta_k)\left(\nabla f_{B_k}(\bm{x}_k) - \nabla f(\bm{x}_{k})\right)\right\rangle \\
& \quad + \beta_k\left(\frac{\epsilon}{2}\left\|\bm{d}_{k-1} - \nabla f(\bm{x}_k)\right\|^2 + \frac{1}{2\epsilon}\left\| \nabla f(\bm{x}_{k}) - \nabla f(\bm{x}_{k+1})\right\|^2\right),
\end{split}
\end{align}
where $\epsilon > 0$, and in the first and second inequality, we used Young's inequality. By the condition (C1)($f$ is $L$--smooth), we have
\begin{align} \label{L_smooth}
\left\| \nabla f(\bm{x}_{k}) - \nabla f(\bm{x}_{k+1})\right\|^2 \leq L^2\|\bm{x}_k - \bm{x}_{k+1}\|^2 = \alpha_k^2L^2\|\bm{m}_k\|^2.
\end{align}

Combining \eqref{L_smooth} with \eqref{d} ensures that
\begin{align*}
\left\| \bm{d}_k - \nabla f(\bm{x}_{k+1}) \right\|^2 &\leq \left(\beta_k^2 + \frac{\beta_k\epsilon}{2}\right)\left\| \bm{d}_{k-1} - \nabla f(\bm{x}_k)\right\|^2 + 2(1-\beta_k)^2\left\|\nabla f_{B_k}(\bm{x}_k) - \nabla f(\bm{x}_{k})\right\|^2 \\
& \quad + 2\alpha_k^2L^2\|\bm{m}_k\|^2 + 2\beta_k\left\langle\bm{d}_{k-1} - \nabla f(\bm{x}_k), (1 - \beta_k)\left(\nabla f_{B_k}(\bm{x}_k) - \nabla f(\bm{x}_{k})\right)\right\rangle \\
& \quad + \frac{\beta_k\alpha_k^2L^2}{2\epsilon}\|\bm{m}_k\|^2 \\
&\leq \left(\beta_k^2 + \frac{\beta_k\epsilon}{2}\right)\left\| \bm{d}_{k-1} - \nabla f(\bm{x}_k)\right\|^2 + 2(1-\beta_k)^2\left\|\nabla f_{B_k}(\bm{x}_k) - \nabla f(\bm{x}_{k})\right\|^2 \\
& \quad + 2\beta_k\left\langle\bm{d}_{k-1} - \nabla f(\bm{x}_k), (1 - \beta_k)\left(\nabla f_{B_k}(\bm{x}_k) - \nabla f(\bm{x}_{k})\right)\right\rangle \\
& \quad + \alpha_k^2L^2\left(2 + \frac{1}{2\epsilon}\right)\|\bm{m}_k\|^2,
\end{align*}
where in the last inequality we used $\beta_k < 1$. Since $\beta_k < 1$, we can choose $\epsilon = 2(1 - \beta_k) > 0$. So, we have
\begin{align} \label{d_2}
\begin{split}
\left\| \bm{d}_k - \nabla f(\bm{x}_{k+1}) \right\|^2 &\leq \beta_k\left\| \bm{d}_{k-1} - \nabla f(\bm{x}_k)\right\|^2 + 2(1-\beta_k)^2\left\|\nabla f_{B_k}(\bm{x}_k) - \nabla f(\bm{x}_{k})\right\|^2 \\
& \quad + 2\beta_k\left\langle\bm{d}_{k-1} - \nabla f(\bm{x}_k), (1 - \beta_k)\left(\nabla f_{B_k}(\bm{x}_k) - \nabla f(\bm{x}_{k})\right)\right\rangle \\
& \quad + \alpha_k^2L^2\left(2 + \frac{1}{1 - \beta_k}\right)\|\bm{m}_k\|^2.
\end{split}
\end{align}

Condition (C2) guarantees that 
\begin{align} \label{estimation_in}
\mathbb{E}_{\xi_k} \left[\nabla f_{B_k} (\bm{x}_k) | \bm{x}_k \right] = \nabla f(\bm{x}_k),
\quad \mathbb{V}_{\xi_k} \left[ \nabla f_{B_k} (\bm{x}_k) | \bm{x}_k \right] \leq \frac{\sigma^2}{b_k}.
\end{align}

Taking the expectation conditioned on $\bm{x}_k$ on both sides of \eqref{d_2} together with \eqref{estimation_in}, guarantees that, for all $k \in \mathbb{N}_0$,
\begin{align}
\begin{split} \label{d-f}
\mathbb{E}\left[\left\| \bm{d}_k - \nabla f(\bm{x}_{k+1}) \right\|^2 | \bm{x}_k \right] &\leq \beta_k\left\| \bm{d}_{k-1} - \nabla f(\bm{x}_k)\right\|^2 + 2(1-\beta_k)^2\frac{\sigma^2}{b_k} \\
& \quad + 2\beta_k\mathbb{E}\left[\left\langle\bm{d}_{k-1} - \nabla f(\bm{x}_k), (1 - \beta_k)\left(\nabla f_{B_k}(\bm{x}_k) - \nabla f(\bm{x}_{k})\right)\right\rangle | \bm{x}_k \right] \\
& \quad + \alpha_k^2\left(2 + \frac{1}{1 - \beta_k}\right)\mathbb{E}\left[\|\bm{m}_k\|^2 | \bm{x}_k\right] \\
&\leq \beta_k\left\| \bm{d}_{k-1} - \nabla f(\bm{x}_k)\right\|^2 + 2(1-\beta_k)^2\frac{\sigma^2}{b_k} \\
&\quad + \alpha_k^2L^2\left(2 + \frac{1}{1 -\beta_k}\right)\mathbb{E}\left[\|\bm{m}_k\|^2 | \bm{x}_k \right].
\end{split}
\end{align}

\endproof

\subsection{Proof of Theorem \ref{th_1}}
\proof
Condition (C1)($f$ is $L$--smooth) implies that the descent lemma holds (see e.g. Lemma 5.7 in \citep{doi:10.1137/1.9781611974997}), i.e. for all $k \in \mathbb{N}_0$,
\begin{align} \label{DL}
f(\bm{x}_{k+1}) &\leq f(\bm{x}_k) + \left\langle \nabla f(\bm{x}_k), \bm{x}_{k+1} - \bm{x}_k\right\rangle + \frac{L}{2}\| \bm{x}_{k+1} - \bm{x}_k \|^2,
\end{align}
which, together with the update rule of mini-batch QHM, implies that
\begin{align} \label{th1_qhm_DL}
f(\bm{x}_{k+1}) &\leq f(\bm{x}_k) - \alpha_k\left\langle \nabla f(\bm{x}_k), \bm{m}_k\right\rangle + \frac{\alpha_k^2L}{2}\left\| \bm{m}_k \right\|^2.
\end{align}

Furthermore, the update rule of mini-batch QHM also ensures that
\begin{align}
\begin{split} \label{m_to_d}
\bm{m}_k
&= (1 - \gamma_k)\nabla f_{B_k}(\bm{x}_k) + \gamma_k\bm{d}_k \\
&= (1 - \gamma_k)\nabla f_{B_k}(\bm{x}_k) + \gamma_k\left((1 - \beta_k)\nabla f_{B_k}(\bm{x}_k) + \beta_k\bm{d}_{k-1}\right) \\
&= (1 - \gamma_k\beta_k)\nabla f_{B_k}(\bm{x}_k) + \gamma_k\beta_k\bm{d}_{k-1}.
\end{split}
\end{align}

Substituting \eqref{m_to_d} for \eqref{th1_qhm_DL}, yields
\begin{align*}
f(\bm{x}_{k+1}) &\leq f(\bm{x}_k) - \alpha_k(1 - \gamma_k\beta_k)\left\langle \nabla f(\bm{x}_k), \nabla f_{B_k}(\bm{x}_k)\right\rangle - \alpha_k\gamma_k\beta_k\left\langle \nabla f(\bm{x}_k), \bm{d}_{k-1}\right\rangle \\
&\quad + \frac{\alpha_k^2L}{2}\left\| (1 - \gamma_k\beta_k)\nabla f_{B_k}(\bm{x}_k) + \gamma_k\beta_k\bm{d}_{k-1} \right\|^2 \\
&= f(\bm{x}_k) - \alpha_k(1 - \gamma_k\beta_k)\left\langle \nabla f(\bm{x}_k), \nabla f_{B_k}(\bm{x}_k)\right\rangle - \alpha_k\gamma_k\beta_k\left\langle \nabla f(\bm{x}_k), \bm{d}_{k-1}\right\rangle \\
&\quad + \frac{\alpha_k^2L}{2}\left((1 - \gamma_k\beta_k)^2\| \nabla f_{B_k}(\bm{x}_k) \|^2 +2(1-\gamma_k\beta_k)\gamma_k\beta_k\left\langle \nabla f_{B_k}(\bm{x}_k), \bm{d}_{k-1}\right\rangle + \gamma_k^2\beta_k^2\| \bm{d}_{k-1} \|^2\right) \\
&\leq f(\bm{x}_k) - \alpha_k(1 - \gamma_k\beta_k)\left\langle \nabla f(\bm{x}_k), \nabla f_{B_k}(\bm{x}_k)\right\rangle + \alpha_k\gamma_k\beta_k\| \nabla f(\bm{x}_k)\|\| \bm{d}_{k-1}\| \\
&\quad + \frac{\alpha_k^2L}{2}\left((1 - \gamma_k\beta_k)^2\| \nabla f_{B_k}(\bm{x}_k) \|^2 + 2(1-\gamma_k\beta_k)\gamma_k\beta_k\| \nabla f_{B_k}(\bm{x}_k)\|\| \bm{d}_{k-1}\| + \gamma_k^2\beta_k^2\| \bm{d}_{k-1} \|^2\right),
\end{align*}
where in the last inequality we used the Cauchy-Schwarz inequality. Using Lemma \ref{bound}, we have
\begin{align}
\begin{split} \label{th1_qhm_DL_2}
f(\bm{x}_{k+1}) &\leq f(\bm{x}_k) - \alpha_k(1 - \gamma_k\beta_k)\left\langle \nabla f(\bm{x}_k), \nabla f_{B_k}(\bm{x}_k)\right\rangle + \alpha_k\gamma_k\beta_kG^2 \\
&\quad + \frac{\alpha_k^2L}{2}\left((1 - \gamma_k\beta_k)^2\| \nabla f_{B_k}(\bm{x}_k) \|^2 + 2(1-\gamma_k\beta_k)\gamma_k\beta_kG^2 + \gamma_k^2\beta_k^2G^2\right) \\
&= f(\bm{x}_k) - \alpha_k(1 - \gamma_k\beta_k)\left\langle \nabla f(\bm{x}_k), \nabla f_{B_k}(\bm{x}_k)\right\rangle + \alpha_k\gamma_k\beta_kG^2 \\
&\quad + \frac{\alpha_k^2L}{2}\left((1 - \gamma_k\beta_k)^2\| \nabla f_{B_k}(\bm{x}_k) \|^2 + (2-\gamma_k\beta_k)\gamma_k\beta_kG^2\right)
\end{split}
\end{align}

From \eqref{estimation_in}, we have
\begin{align}
\begin{split} \label{estimation_norm}
\mathbb{E}_{\xi_k} \left[\|\nabla f_{B_k} (\bm{x}_k)\|^2 | \bm{x}_k \right] &= \| \nabla f(\bm{x}_k) \|^2 + \mathbb{V}_{\xi_k}\left[\nabla f_{B_k} (\bm{x}_k)|\bm{x}_k\right] \\
&\leq \| \nabla f(\bm{x}_k) \|^2 + \frac{\sigma^2}{b_k}.
\end{split}
\end{align}

Taking the expectation conditioned on $\bm{x}_k$ on both sides of \eqref{th1_qhm_DL_2}, together with \eqref{estimation_in} and \eqref{estimation_norm}, guarantees that, for all $k \in \mathbb{N}_0$,
\begin{align*}
\mathbb{E}\left[f(\bm{x}_{k+1}) | \bm{x}_k \right] &\leq f(\bm{x}_k) - \alpha_k(1 - \gamma_k\beta_k)\| \nabla f(\bm{x}_k)\|^2 + \alpha_k\gamma_k\beta_kG^2 \\
&\quad + \frac{\alpha_k^2 L}{2}\left((1 - \gamma_k\beta_k)^2\left(\frac{\sigma^2}{b_k} + \| \nabla f(\bm{x}_k) \|^2\right) + (2-\gamma_k\beta_k)\gamma_k\beta_kG^2\right) \\
&\leq f(\bm{x}_k) -\frac{1}{2}\alpha_k(1 - \gamma_k\beta_k)(2 - \alpha_kL(1 - \gamma_k\beta_k))\| \nabla f(\bm{x}_k)\|^2 + \frac{L\sigma^2}{2}\frac{\alpha_k^2(1-\gamma_k\beta_k)^2}{b_k} \\
&\quad +\frac{1}{2}\alpha_k\gamma_k\beta_k(2 + \alpha_kL(2-\gamma_k\beta_k))G^2.
\end{align*}

Therefore, taking the total expectation on both sides of the above inequality ensures that, for all $k \in \mathbb{N}_0$,
\begin{align*}
\frac{1}{2}\alpha_k(1 - \gamma_k\beta_k)(2 - \alpha_k L(1 - \gamma_k\beta_k))\mathbb{E}\left[\| \nabla f(\bm{x}_k)\|^2\right] &\leq \mathbb{E}\left[f(\bm{x}_k) - f(\bm{x}_{k+1})\right] +\frac{L\sigma^2}{2}\frac{\alpha_k^2(1-\gamma_k\beta_k)^2}{b_k} \\
&\quad +\frac{G^2}{2}\alpha_k\gamma_k\beta_k(2 + \alpha_kL(2-\gamma_k\beta_k)).
\end{align*}

Let $K \geq 1$. Summing the above inequality from $k = 0$ to $k = K - 1$ ensures that
\begin{align*}
\frac{1}{2}\sum_{k=0}^{K-1}\alpha_k(1 - \gamma_k\beta_k)(2 - \alpha_k L(1 - \gamma_k\beta_k))\mathbb{E}\left[\| \nabla f(\bm{x}_k)\|^2\right] &\leq \mathbb{E}\left[f(\bm{x}_0) - f(\bm{x}_{K})\right] +\frac{L\sigma^2}{2}\sum_{k=0}^{K-1}\frac{\alpha_k^2(1-\gamma_k\beta_k)^2}{b_k} \\
&\quad +\frac{G^2}{2}\sum_{k=0}^{K-1}\alpha_k\gamma_k\beta_k(2 + \alpha_kL(2-\gamma_k\beta_k)).
\end{align*}

From (C1)(the lower bound $f_\star$ of $f$), $\alpha_k \in [\alpha_{\min}, \alpha_{\max}]$, and $0 \leq \gamma_k\beta_k \leq \overline{\gamma\beta} < 1$, we have that
\begin{align*}
\frac{1}{2}(1 - \overline{\gamma\beta})(2 - \alpha_{\max}L)\sum_{k=0}^{K-1}\alpha_k\mathbb{E}\left[\| \nabla f(\bm{x}_k)\|^2\right] &\leq f(\bm{x}_0) - f_\star +\frac{L\sigma^2}{2}\sum_{k=0}^{K-1}\frac{\alpha_k^2}{b_k} \\
&\quad + \alpha_{\max}(1 + \alpha_{\max}L)G^2\sum_{k=0}^{K-1}\gamma_k\beta_k.
\end{align*}

Since, $\alpha_{\max} \leq \frac{2}{L}$, we obtain
\begin{align*}
\min_{k \in [0:K-1]}\mathbb{E}\left[\left\|\nabla f(\bm{x}_k)\right\|^2\right] &\leq \frac{2(f(\bm{x}_0) - f_\star)}{(1-\overline{\gamma\beta})(2-\alpha_{\max}L)}\underbrace{\frac{1}{\sum_{k=0}^{K-1}\alpha_k}}_{A_K} \\
&\quad + \frac{L\sigma^2}{(1-\overline{\gamma\beta})(2-\alpha_{\max}L)}\underbrace{\frac{\sum_{k=0}^{K-1}\alpha_k^2/b_k}{\sum_{k=0}^{K-1}\alpha_k}}_{B_K} \\
&\quad + \frac{2\alpha_{\max}(1+\alpha_{\max}L)G^2}{(1-\overline{\gamma\beta})(2-\alpha_{\max}L)}\underbrace{\frac{\sum_{k=0}^{K-1}\gamma_k\beta_k}{\sum_{k=0}^{K-1}\alpha_k}}_{C_k}.
\end{align*}

This proves Theorem \ref{th_1}(i). Second, we assume the condition of Theorem \ref{th_1}(ii). $\sum_{k=0}^{+\infty}\alpha_k = +\infty$, $\sum_{k=0}^{+\infty}\gamma_k\beta_k < +\infty$, and $\sum_{k=0}^{+\infty}\frac{\alpha^2_k}{b_k} < +\infty$ imply that 
\begin{align*}
\lim_{K \to \infty} A_K = \lim_{K \to \infty} B_K = \lim_{K \to \infty} C_K = 0.
\end{align*}

This proves Theorem \ref{th_1}(ii).
\endproof

\subsection{Proof of Corollary \ref{cor_1}}
\proof

[Constant LR]: We have that
\begin{align*}
A_K = \frac{1}{\sum_{k=0}^{K-1}\alpha_k} = \frac{1}{\alpha_{\max}K}.
\end{align*}

In the case of using [Constant BS \eqref{constant_batch}], we obtain 
\begin{align*}
B_K = \frac{\sum_{k=0}^{K-1}\alpha_k^2}{b\sum_{k=0}^{K-1}\alpha_k} = \frac{\alpha_{\max}}{b}.
\end{align*}

Also, in the case of using [Exponential BS \eqref{exp_batch}], since $b'_m$ is monotone increasing, $T_m = \left\lceil\frac{n}{b'_m}\right\rceil$ is monotone decreasing. Hence, we have that
\begin{align*}
\sum_{k=0}^{K-1}\frac{1}{b_k} = \sum_{m=0}^{M-1}\sum_{t=0}^{T_m-1}\frac{1}{b'_m} \leq T_0\sum_{m=0}^{M-1}\frac{1}{b'_m} = \frac{T_0E}{b_0}\sum_{m'=0}^{M'}\frac{1}{\delta^{m'}} \leq \frac{T_0E\delta}{b_0(\delta-1)}.
\end{align*}

Therefore, 
\begin{align*}
B_K = \frac{\sum_{k=0}^{K-1}\alpha_k^2/b_k}{\sum_{k=0}^{K-1}\alpha_k} \leq \frac{\alpha_{\max}T_0E\delta}{b_0(\delta-1)K}.
\end{align*}

In the case of [Exponential BS \eqref{exp_batch}], when using [Step Decay Beta \eqref{step_decay_beta}] and [Step Decay Gamma \eqref{step_decay_gamma}], we find that
\begin{align*}
\sum_{k=0}^{K-1}\gamma_k\beta_k = \sum_{m=0}^{M-1}\sum_{t=0}^{T_m-1}\gamma'_m\beta'_m \leq \gamma_{\max}\beta_{\max}T_0E\sum_{m'=0}^{M'}(\lambda\zeta)^{m'} \leq \frac{\gamma_{\max}\beta_{\max}T_0E}{1-\lambda\zeta}.
\end{align*}

This result shows that, by setting $T_m = T_0 = \left\lceil\frac{n}{b}\right\rceil$, we can obtain the same outcome even when using [Constant BS \eqref{constant_batch}].

[Decaying Squared LR]: We have that
\begin{align*}
\alpha_k = \frac{\alpha_{\max}}{\sqrt{\left\lfloor\frac{k}{T_m}\right\rfloor+1}} \geq \frac{\alpha_{\max}}{\sqrt{k+1}}.
\end{align*}

Therefore,
\begin{align*}
\sum_{k=0}^{K-1}\alpha_k \geq \alpha_{\max}\int_{0}^{K}\frac{dk}{\sqrt{k+1}} = 2\alpha_{\max}(\sqrt{K+1}-1).
\end{align*}

Hence,
\begin{align*}
&A_K = \frac{1}{\sum_{k=0}^{K-1}\alpha_k} \leq \frac{1}{2\alpha_{\max}(\sqrt{K+1}-1)}, \\
&C_K = \frac{\sum_{k=0}^{K-1}\gamma_k\beta_k}{\sum_{k=0}^{K-1}\alpha_k} \leq \frac{\gamma_{\max}\beta_{\max}T_0E}{2\alpha_{\max}(1-\lambda\zeta)(\sqrt{K+1}-1)}.
\end{align*}

We also have
\begin{align*}
\sum_{m=0}^{M-1}{\alpha'}_k^2 \leq \alpha_{\max}^2\left(1 + \int_0^M\frac{dm}{m+1}\right) = \alpha_{\max}^2\left(1+\log(M+1)\right) \leq \alpha_{\max}^2\left(1+\log(K+1)\right),
\end{align*}
which implies that
\begin{align*}
\sum_{k=0}^{K-1}\alpha_k^2 = T\sum_{m=0}^{M-1}{\alpha'}_k^2 \leq \alpha_{\max}^2T\left(1+\log(K+1)\right).
\end{align*}

Therefore, in the case of [Constant BS \eqref{constant_batch}], 
\begin{align*}
B_K = \frac{\sum_{k=0}^{K-1}\alpha_k^2}{b\sum_{k=0}^{K-1}\alpha_k} \leq \frac{\alpha_{\max}T\left(1+\log(K+1)\right)}{2b(\sqrt{K+1}-1)}.
\end{align*}

Similarly, in the case of [Exponential BS \eqref{exp_batch}],
\begin{align*}
B_K = \frac{\sum_{k=0}^{K-1}\alpha_k^2/b_k}{\sum_{k=0}^{K-1}\alpha_k} \leq \frac{\alpha_{\max}T_0E\delta}{2b_0(\delta-1)(\sqrt{K+1}-1)}.
\end{align*}

[Cosine LR]: If $M=1$, [Cosine LR] corresponds to [Constant LR]. So, let $M \geq 2$. And let $T_{\min}$ satisfy $0 \leq T_{\min} \leq T_m$. Consequently, we have 
\begin{align} \label{cos_ane}
\sum_{k=0}^{K-1}\alpha_k &= \alpha_{\min}K + \frac{\alpha_{\max}-\alpha_{\min}}{2}K + \frac{\alpha_{\max}-\alpha_{\min}}{2}\sum_{m=0}^{M-1}T_m\cos\frac{m\pi}{M-1} \\
&\geq \alpha_{\min}K + \frac{\alpha_{\max}-\alpha_{\min}}{2}K + \frac{\alpha_{\max}-\alpha_{\min}}{2}T_{\min}\sum_{m=0}^{M-1}\cos\frac{m\pi}{M-1}.
\end{align}

Here, $\sum_{m=0}^{M-1}\cos\frac{m\pi}{M-1} = 0$ holds. Actually, in the case of $M = 2F(F\in\mathbb{N})$, we have that
\begin{align} \label{cos_sum}
\sum_{m=0}^{2F-1}\cos\frac{m\pi}{M-1} = \cos0+\cos\frac{\pi}{M-1}+\cdots+\cos\frac{F-1}{M-1}\pi + \cos\frac{F}{M-1}\pi+\cdots+\cos\frac{M-2}{M-1}\pi+\cos\pi.
\end{align}

From $\cos(\pi-x)=-\cos x$, \eqref{cos_sum} satisfies
\begin{align*}
\cos 0+\cos\pi = \cos\frac{\pi}{M-1}+\cos\frac{M-2}{M-1}\pi = \cdots = \cos\frac{k-1}{M-1}\pi+\cos\frac{k}{M-1}\pi = 0.
\end{align*}

Therefore, 
\begin{align*}
\sum_{m=0}^{M-1}\cos\frac{m\pi}{M-1} = 0.
\end{align*}

Next, in the case of $M = 2F+1$, we have that
\begin{align*}
\sum_{m=0}^{2F}\cos\frac{m\pi}{M-1} = \cos0+\cos\frac{\pi}{M-1}+\cdots+\cos\frac{F-1}{M-1}\pi + \cos\frac{F}{M-1}\pi+\cos\frac{F+1}{M-1}\pi+\cdots+\cos\pi.
\end{align*}

Since $\cos\frac{G}{M-1}\pi = \cos\frac{k}{2k}\pi=0$ and $\cos(\pi-x)=-\cos x$, the same holds as in the case above. Hence, $\sum_{m=0}^{M-1}\cos\frac{m\pi}{M-1} = 0$ holds. So, \eqref{cos_ane} satisfies
\begin{align*}
\sum_{k=0}^{K-1}\alpha_k \geq \frac{\alpha_{\max}+\alpha_{\min}}{2}K.
\end{align*}

Therefore,
\begin{align*}
&A_K = \frac{1}{\sum_{k=0}^{K-1}\alpha_k} \leq \frac{2}{(\alpha_{\max}+\alpha_{\min})K}, \\
&C_K = \frac{\sum_{k=0}^{K-1}\gamma_k\beta_k}{\sum_{k=0}^{K-1}\alpha_k} \leq \frac{2\gamma_{\max}\beta_{\max}T_0E}{(1-\lambda\zeta)(\alpha_{\max}+\alpha_{\min})K}.
\end{align*}

In the case of using [Constant BS \eqref{constant_batch}], we have that
\begin{align*}
\sum_{k=0}^{K-1}\alpha_k^2 &= \alpha_{\min}^2K+\alpha_{\min}(\alpha_{\max}-\alpha_{\min})K+\alpha_{\min}(\alpha_{\max}-\alpha_{\min})T_0\sum_{m=0}^{M-1}\cos\frac{m\pi}{M-1} \\
&\quad + \frac{(\alpha_{\max}-\alpha_{\min})^2}{4}K+\frac{(\alpha_{\max}-\alpha_{\min})^2}{2}T_0\sum_{m=0}^{M-1}\cos\frac{m\pi}{M-1}+\frac{(\alpha_{\max}-\alpha_{\min})^2}{4}T_0\sum_{m=0}^{M-1}\cos^2\frac{m\pi}{M-1} \\
\end{align*}

Here, 
\begin{align*}
\sum_{m=0}^{M-1}\cos^2\frac{m\pi}{M-1} &\leq 1 + \int_{0}^{M-1}\cos^2\frac{m\pi}{M-1}dm \\
&= 1+\frac{1}{2}\int_{0}^{M-1}1+\cos2\frac{m\pi}{M-1}dm \\
&=\frac{M+1}{2}.
\end{align*}

From this and $\sum_{m=0}^{M-1}\cos\frac{m\pi}{M-1} = 0$, we obtain
\begin{align*}
\sum_{k=0}^{K-1}\alpha_k^2 &\leq \alpha_{\min}\alpha_{\max}K + \frac{\alpha_{\max}^2}{4}K -\frac{\alpha_{\max}\alpha_{\min}}{2}K + \frac{\alpha_{\min}^2}{4}K + \frac{(M+1)(\alpha_{\max}-\alpha_{\min})^2}{4}T_0 \\
&= \left(\frac{\alpha_{\max}+\alpha_{\min}}{2}\right)^2K + \frac{(M+1)(\alpha_{\max}-\alpha_{\min})^2}{4}T_0 \\
&= \left(\left(\frac{\alpha_{\max}+\alpha_{\min}}{2}\right)^2 + \frac{(\alpha_{\max}-\alpha_{\min})^2}{4}\right)K + \frac{(\alpha_{\max}-\alpha_{\min})^2}{4}T_0 \\
&= \frac{\alpha_{\max}^2+\alpha_{\min}^2}{2}K + \frac{(\alpha_{\max}-\alpha_{\min})^2}{4}T_0 \\
&\leq \frac{(\alpha_{\max}+\alpha_{\min})^2}{2}K + \frac{(\alpha_{\max}-\alpha_{\min})^2}{4}T_0,
\end{align*}
which implies that
\begin{align*}
B_K = \frac{\sum_{k=0}^{K-1}\alpha_k^2}{b\sum_{k=0}^{K-1}\alpha_k} \leq \frac{\alpha_{\max}+\alpha_{\min}}{b}+\frac{T_0(\alpha_{\max}-\alpha_{\min})^2}{2b(\alpha_{\max}+\alpha_{\min})K}.
\end{align*}

On the other hand, in the case of using [Exponential BS \eqref{exp_batch}], we obtain 
\begin{align*}
B_K = \frac{\sum_{k=0}^{K-1}\alpha_k^2/b_k}{\sum_{k=0}^{K-1}\alpha_k} \leq \frac{\alpha_{\max}^2\sum_{k=0}^{K-1}1/b_k}{\sum_{k=0}^{K-1}\alpha_k} \leq \frac{2\alpha_{\max}^2T_0E\delta}{b_0(\delta-1)(\alpha_{\max}+\alpha_{\min})K}.
\end{align*}

[Polynomial LR]: Since $f(x)=(1-x)^p$ is monotone decreasing for $x \in [0,1]$, we have that
\begin{align*}
\int_{0}^{1}(1-x)^pdx \leq \frac{1}{M}\sum_{m=0}^{M-1}\left(1-\frac{m}{M}\right)^p,
\end{align*}
which implies that
\begin{align} \label{p_sum}
M\int_{0}^{1}(1-x)^pdx \leq \sum_{m=0}^{M-1}\left(1-\frac{m}{M-1}\right)^p. 
\end{align}

Since $\int_{0}^{1}(1-x)^pdx = \frac{1}{p+1}$, \eqref{p_sum} satisfies
\begin{align*}
\sum_{m=0}^{M-1}\left(1-\frac{m}{M-1}\right)^p \geq \frac{M}{p+1}.
\end{align*}

Therefore, in the case of using [Constant BS \eqref{constant_batch}],
\begin{align*}
\sum_{k=0}^{K-1}\alpha_k &= \alpha_{\min}K + (\alpha_{\max}-\alpha_{\min})T_0\sum_{m=0}^{M-1}\left(1-\frac{m}{M}\right)^p \\
&\geq \alpha_{\min}K + (\alpha_{\max}-\alpha_{\min})\frac{T_0M}{p+1} \\
&= \frac{\alpha_{\max}+p\alpha_{\min}}{p+1}K,
\end{align*}
which implies that
\begin{align*}
&A_K = \frac{1}{\sum_{k=0}^{K-1}\alpha_k} \leq \frac{p+1}{(\alpha_{\max}+p\alpha_{\min})K}, \\
&C_K = \frac{\sum_{k=0}^{K-1}\gamma_k\beta_k}{\sum_{k=0}^{K-1}\alpha_k} \leq \frac{\gamma_{\max}\beta_{\max}(p+1)T_0E}{(\alpha_{\max}+p\alpha_{\min})(1-\lambda\zeta)K}.
\end{align*}

On the other hand, since $f(x)=(1-x)^p$ and $g(x)=(1-x)^{2p}$ are monotone decreasing for $x \in [0,1)$, we have 
\begin{align*}
\frac{1}{M}\sum_{m=0}^{M-1}\left(1-\frac{m}{M}\right)^p \leq \frac{1}{M} + \int_{0}^{1}(1-x)^pdx, \quad \frac{1}{M}\sum_{m=0}^{M-1}\left(1-\frac{m}{M}\right)^{2p} \leq \frac{1}{M} + \int_{0}^{1}(1-x)^{2p}dx,
\end{align*}
which implies that
\begin{align} \label{p_sum_2}
\sum_{m=0}^{M-1}\left(1-\frac{m}{M}\right)^p \leq 1 + M\int_{0}^{1}(1-x)^pdx, \quad \sum_{m=0}^{M-1}\left(1-\frac{m}{M}\right)^{2p} \leq 1 + M\int_{0}^{1}(1-x)^{2p}dx.
\end{align}

Since $\int_{0}^{1}(1-x)^pdx = \frac{1}{p+1}$ and $\int_{0}^{1}(1-x)^{2p}dx = \frac{1}{2p+1}$, \eqref{p_sum_2} satisfies 
\begin{align*}
\sum_{m=0}^{M-1}\left(1-\frac{m}{M}\right)^p \leq 1 + \frac{M}{p+1}, \quad \sum_{m=0}^{M-1}\left(1-\frac{m}{M}\right)^{2p} \leq 1 + \frac{M}{2p+1}.
\end{align*}

Hence,
\begin{align*}
\sum_{k=0}^{K-1}\alpha_k^2 &= \alpha_{\min}^2K + 2\alpha_{\min}(\alpha_{\max}-\alpha_{\min})T_0\sum_{m=0}^{M-1}\left(1-\frac{m}{M}\right)^p + (\alpha_{\max}-\alpha_{\min})^2T_0\sum_{m=0}^{M-1}\left(1-\frac{m}{M}\right)^{2p} \\
&\leq \alpha_{\min}^2K + 2\alpha_{\min}(\alpha_{\max}-\alpha_{\min})T_0\left(1 + \frac{M}{p+1}\right) + (\alpha_{\max}-\alpha_{\min})^2T_0\left(1 + \frac{M}{2p+1}\right) \\
&= \frac{(p+1)\alpha_{\max}^2+2p\alpha_{\max}\alpha_{\min}+2p^2\alpha_{\min}^2}{(p+1)(2p+1)}K + (\alpha_{\max}^2-\alpha_{\min}^2)T_0,
\end{align*}
which implies that
\begin{align*}
B_K = \frac{\sum_{k=0}^{K-1}\alpha_k^2}{b\sum_{k=0}^{K-1}\alpha_k} \leq \frac{(p+1)\alpha_{\max}^2+2p\alpha_{\max}\alpha_{\min}+2p^2\alpha_{\min}^2}{b(2p+1)(\alpha_{\max}+p\alpha_{\min})} + \frac{(p+1)(\alpha_{\max}^2-\alpha_{\min}^2)T_0}{(\alpha_{\max}+p\alpha_{\min})K}.
\end{align*}

In the case of using [Exponential BS \eqref{exp_batch}], the following inequality holds for all $m \in [0:M-1]$ and $k \in [0:K-1]$:
\begin{align} \label{poly_ine}
\frac{\lfloor\frac{k}{T_m}\rfloor}{M} \leq \frac{k}{K}.
\end{align}

Actually, let $S_m = \sum_{m'=0}^{m}T_{m'}$. Since $\lfloor\frac{k}{T_m}\rfloor = m$ for all $k \in [S_m:S_{m+1}-1]$, showing that the following inequality holds for all $m \in [0:M-1]$ is sufficient to establish \eqref{poly_ine}:
\begin{align*}
\frac{m}{M} \leq \frac{S_m}{K}.
\end{align*}

From the definition of $S_m$ and $K$,
\begin{align*}
M\sum_{m'=0}^{m}T_{m'} - m\sum_{m'=0}^{M-1}T_{m'} &= (M-m)\sum_{m'=0}^{m}T_{m'}-m\sum_{m'=m+1}^{M-1}T_{m'} \\
&\geq ((M-m)(m+1) -m(M-1-(m+1)+1))T_m \\
&= MT_m \geq 0,
\end{align*}
where the first inequality holds because ${T_m}$ is monotone decreasing. Since f(x) is monotone decreasing, from using \eqref{poly_ine}, we obtain 
\begin{align*}
\alpha_k &= \alpha_{\min} + (\alpha_{\max}-\alpha_{\min})\left(1-\frac{\lfloor\frac{k}{T_m}\rfloor}{M}\right)^p \\
&\geq \alpha_{\min} + (\alpha_{\max}-\alpha_{\min})\left(1-\frac{k}{K}\right)^p,
\end{align*}
which implies that
\begin{align*}
\sum_{k=0}^{K-1}\alpha_k &\geq \alpha_{\min}K + (\alpha_{\max}-\alpha_{\min})\sum_{k=0}^{K-1}\left(1-\frac{k}{K}\right)^p \\
&= \alpha_{\min}K + \frac{\alpha_{\max}-\alpha_{\min}}{p+1}K \\
&= \frac{\alpha_{\max}-p\alpha_{\min}}{p+1}K.
\end{align*}

As a result, we have that
\begin{align*}
&A_K = \frac{1}{\sum_{k=0}^{K-1}\alpha_k} \leq \frac{p+1}{(\alpha_{\max}+p\alpha_{\min})K}, \\
&B_K = \frac{\sum_{k=0}^{K-1}\alpha_k^2/b_k}{\sum_{k=0}^{K-1}\alpha_k} \leq \frac{\alpha_{\max}^2\sum_{k=0}^{K-1}1/b_k}{\sum_{k=0}^{K-1}\alpha_k} \leq \frac{\alpha_{\max}^2(p+1)T_{0}E\delta}{b_0(\alpha_{\max}+p\alpha_{\min})(\delta-1)K}. 
\end{align*}

\endproof

\subsection{Proof of Theorem \ref{th_2}}
\proof

From \eqref{th1_qhm_DL}, we have that, for all $k \in \mathbb{N}_0$,
\begin{align}
\begin{split} \label{th2_qhm_DL}
f(\bm{x}_{k+1}) &\leq f(\bm{x}_k) - \alpha_k\left\langle \nabla f(\bm{x}_k), \bm{m}_k\right\rangle + \frac{\alpha_k^2L}{2}\left\| \bm{m}_k \right\|^2 \\
&= f(\bm{x}_k) - \alpha_k\| \nabla f(\bm{x}_k) \|^2 - \alpha_k\left\langle \nabla f(\bm{x}_k), \bm{m}_k - \nabla f(\bm{x}_k)\right\rangle + \frac{\alpha_k^2L}{2}\left\| \bm{m}_k \right\|^2
\end{split}
\end{align}

Here, from \eqref{m_to_d}, we find that
\begin{align}
\begin{split} \label{m-f}
\bm{m}_k - \nabla f(\bm{x}_k) &= (1 - \gamma_k\beta_k)\nabla f_{B_k}(\bm{x}_k) + \gamma_k\beta_k\bm{d}_{k-1} - \nabla f(\bm{x}_k) \\
&= (1 - \gamma_k\beta_k)\left(\nabla f_{B_k}(\bm{x}_k) - \nabla f(\bm{x}_k)\right) + \gamma_k\beta_k\left(\bm{d}_{k-1} - \nabla f(\bm{x}_k)\right).
\end{split}
\end{align}

Substituting \eqref{m-f} for \eqref{th2_qhm_DL} yields
\begin{align}
\begin{split} \label{th2_qhm_DL_2}
f(\bm{x}_{k+1}) &\leq f(\bm{x}_k) - \alpha_k \left\|\nabla f(\bm{x}_k)\right\|^2 - \alpha_k\gamma_k\beta_k\left\langle \nabla f(\bm{x}_k), \bm{d}_{k-1} - \nabla f(\bm{x}_k)\right\rangle \\
& \quad - \alpha_k(1 - \gamma_k\beta_k)\left\langle \nabla f(\bm{x}_k), \nabla f_{B_k}(\bm{x}_k) - \nabla f(\bm{x}_k)\right\rangle + \frac{\alpha_k^2L}{2}\left\| \bm{m}_k \right\|^2 \\
& \leq f(\bm{x}_k) - \alpha_k \left\|\nabla f(\bm{x}_k)\right\|^2 + \alpha_k\gamma_k\beta_k\left( \frac{1}{2}\|\nabla f(\bm{x}_k)\|^2 + \frac{1}{2}\| \bm{d}_{k-1} -\nabla f(\bm{x}_k)\|^2 - \frac{1}{2}\|\bm{d}_{k-1}\|^2 \right) \\
& \quad - \alpha_k(1 - \gamma_k\beta_k)\left\langle \nabla f(\bm{x}_k), \nabla f_{B_k}(\bm{x}_k) - \nabla f(\bm{x}_k)\right\rangle + \frac{\alpha_k^2L}{2}\left\| \bm{m}_k \right\|^2 \\
& \leq f(\bm{x}_k) - \frac{1}{2}\alpha_k \left\|\nabla f(\bm{x}_k)\right\|^2 + \frac{1}{2}\alpha_k\gamma_k\beta_k\left\| \bm{d}_{k-1} - \nabla f(\bm{x}_k)\right\|^2 \\
& \quad - \alpha_k(1 - \gamma_k\beta_k)\left\langle \nabla f(\bm{x}_k), \nabla f_{B_k}(\bm{x}_k) - \nabla f(\bm{x}_k)\right\rangle + \frac{\alpha_k^2L}{2}\left\| \bm{m}_k \right\|^2. \\
\end{split}
\end{align}
where, in the second inequality, we used $2\langle\bm{x},\bm{y}\rangle = \|\bm{x} - \bm{y}\|^2 - \|\bm{x}\|^2 - \|\bm{y}\|^2$ for all $\bm{x},\bm{y} \in \mathbb{R}^d$, and in the last inequality, we used $0 \leq \gamma_k\beta_k \leq 1$. Taking the expectation conditioned on $\bm{x}_k$ on both sides of \eqref{th2_qhm_DL_2} together with \eqref{estimation_in}, guarantees that, for all $k \in \mathbb{N}_0$,
\begin{align}
\begin{split} \label{last_DL}
\mathbb{E}\left[f(\bm{x}_{k+1}) | \bm{x}_k\right] & \leq f(\bm{x}_k) - \frac{1}{2}\alpha_k \left\|\nabla f(\bm{x}_k)\right\|^2 + \frac{1}{2}\alpha_k\gamma_k\beta_k\left\| \bm{d}_{k-1} - \nabla f(\bm{x}_k)\right\|^2 \\
& \quad - \alpha_k(1 - \gamma_k\beta_k)\mathbb{E}\left[\left\langle \nabla f(\bm{x}_k), \nabla f_{B_k}(\bm{x}_k) - \nabla f(\bm{x}_k)\right\rangle | \bm{x}_k\right] + \frac{\alpha_k^2L}{2}\mathbb{E}\left[\left\| \bm{m}_k \right\|^2 | \bm{x}_k\right] \\
& = f(\bm{x}_k) - \frac{1}{2}\alpha_k \left\|\nabla f(\bm{x}_k)\right\|^2 + \frac{1}{2}\alpha_k\gamma_k\beta_k\left\| \bm{d}_{k-1} - \nabla f(\bm{x}_k)\right\|^2 + \frac{\alpha_k^2L}{2}\mathbb{E}\left[\left\| \bm{m}_k \right\|^2 | \bm{x}_k\right]. \\
\end{split}
\end{align}

Adding both sides of \eqref{last_DL} and Lemma \ref{lem_2}, we obtain, from $L \leq L^2$, that
\begin{align}
\begin{split} \label{d-f_DL}
\mathbb{E}\left[f(\bm{x}_{k+1}) + \left\| \bm{d}_k - \nabla f(\bm{x}_{k+1}) \right\|^2 | \bm{x}_k\right]
&\leq f(\bm{x}_k) + \beta_k(\frac{\alpha_k\gamma_k}{2} + 1)\left\| \bm{d}_{k-1} - \nabla f(\bm{x}_k)\right\|^2 - \frac{1}{2}\alpha_k\left\|\nabla f(\bm{x}_k)\right\|^2 \\
&\quad + 2(1-\beta_k)\frac{\sigma^2}{b_k} + \alpha_k^2L^2\left(\frac{5}{2} + \frac{1}{1-\beta_k}\right)\mathbb{E}_k\left[\|\bm{m}_k\|^2 | \bm{x}_k\right].
\end{split}
\end{align}

By using Lemma \eqref{bound} and $\beta_k(\frac{\alpha_k}{2} + 1) \leq 1$ for all $k \in \mathbb{N}_0$, we obtain
\begin{align} \label{DL+(d-f)}
\begin{split}
\mathbb{E}_k\left[f(\bm{x}_{k+1}) + \left\| \bm{d}_k - \nabla f(\bm{x}_{k+1}) \right\|^2\right]
&\leq f(\bm{x}_k) + \left\| \bm{d}_{k-1} - \nabla f(\bm{x}_k)\right\|^2 - \frac{1}{2}\alpha_k\left\|\nabla f(\bm{x}_k)\right\|^2 \\
&\quad + 2(1-\beta_k)\frac{\sigma^2}{b_k} + \alpha_k^2L^2\left(\frac{5}{2} + \frac{1}{1-\beta_k}\right)G^2.
\end{split}
\end{align}

Therefore, taking the total expectation on both sides of the above inequality ensures that, for all $k \in \mathbb{N}_0$,
\begin{align*}
\frac{1}{2}\alpha_k\mathbb{E}\left[\left\|\nabla f(\bm{x}_k)\right\|^2\right]
&\leq \mathbb{E}\left[f(\bm{x}_k) + \left\| \bm{d}_{k-1} - \nabla f(\bm{x}_k)\right\|^2 - f(\bm{x}_{k+1}) + \left\| \bm{d}_k - \nabla f(\bm{x}_{k+1}) \right\|^2\right] \\
&\quad + 2(1-\beta_k)\frac{\sigma^2}{b_k} + \alpha_k^2L^2\left(\frac{5}{2} + \frac{1}{1-\beta_k}\right)G^2
\end{align*}

Moreover, summing from $k = 0$ to $k = K - 1$ ensures that
\begin{align*}
\frac{1}{2}\sum_{k=0}^{K-1}\alpha_k\mathbb{E}\left[\left\|\nabla f(\bm{x}_k)\right\|^2\right]
&\leq \mathbb{E}\left[f(\bm{x}_0) + \left\| \bm{d}_{-1} - \nabla f(\bm{x}_0)\right\|^2 - f(\bm{x}_{K}) + \left\| \bm{d}_{K-1} - \nabla f(\bm{x}_{K}) \right\|^2\right] \\
&\quad + 2\sigma^2\sum_{k=0}^{K-1}\frac{1-\beta_k}{b_k} + \frac{5}{2}L^2G^2\sum_{k=0}^{K-1}\alpha_k^2 + L^2G^2\sum_{k=0}^{K-1}\frac{\alpha_k^2}{1-\beta_k}.
\end{align*}

By setting $\bm{d}_{-1} = \left\| \nabla f(\bm{x}_0)\right\|^2$, from the condition (C1)(the lower bound $f_\star$ of $f$), we find that
\begin{align*}
\frac{1}{2}\sum_{k=0}^{K-1}\alpha_k\mathbb{E}\left[\left\|\nabla f(\bm{x}_k)\right\|^2\right]
&\leq f(\bm{x}_0) - f_\star + 2\sigma^2\sum_{k=0}^{K-1}\frac{1-\beta_k}{b_k} \\
&\quad + \frac{5}{2}L^2G^2\sum_{k=0}^{K-1}\alpha_k^2 + L^2G^2\sum_{k=0}^{K-1}\frac{\alpha_k^2}{1-\beta_k}.
\end{align*}

Therefore,
\begin{align*}
\min_{k \in [0:K-1]}\mathbb{E}\left[\left\|\nabla f(\bm{x}_k)\right\|^2\right]
&\leq 2(f(\bm{x}_0) - f_\star)\underbrace{\frac{1}{\sum_{k=0}^{K-1}\alpha_k}}_{A_K} + 4\sigma^2\underbrace{\sum_{k=0}^{K-1}\frac{1-\beta_k}{b_k}\frac{1}{\sum_{k=0}^{K-1}\alpha_k}}_{B'_K} \\
&\quad + 5L^2G^2\underbrace{\frac{\sum_{k=0}^{K-1}\alpha_k^2}{\sum_{k=0}^{K-1}\alpha_k}}_{C'_K} + 2L^2G^2\underbrace{\sum_{k=0}^{K-1}\frac{\alpha_k^2}{1-\beta_k}\frac{1}{\sum_{k=0}^{K-1}\alpha_k}}_{D'_K}.
\end{align*}

This proves Theorem \ref{th_2}(i). Second, we assume the condition of Theorem \ref{th_2}(ii). From $\sum_{k=0}^{+\infty}\alpha_k = +\infty$, $\sum_{k=0}^{\infty}\frac{1-\beta_k}{b_k} < +\infty$, and $\sum_{k=0}^{+\infty}\frac{\alpha_k^2}{1-\beta_k} < +\infty$, we find that
\begin{align*}
\lim_{K \to \infty} A_K = \lim_{K \to \infty} B'_K = \lim_{K \to \infty} C'_K = \lim_{K \to \infty} D'_K = 0.
\end{align*}

This proves Theorem \ref{th_2}(ii). 
\endproof

\subsection{Proof of Corollary \ref{cor_2}}
\proof

[Decaying Squared LR] $\land$ [Constant Beta]: The upper bounds of $A'_K$ and $C'_K$ are provided in Corollary \ref{cor_1}. So, we will only provide upper bounds of $B'_K$ and $D'_K$. In particular, we have that
\begin{align*}
\sum_{k=0}^{K-1}(1-\beta_k) = K(1-\beta_{\max}).
\end{align*}

Therefore, in the case of using [Constant BS \eqref{constant_batch}], 
\begin{align*}
B'_K = \frac{\sum_{k=0}^{K-1}(1-\beta_k)}{b\sum_{k=0}^{K-1}\alpha_k} = \frac{K(1-\beta_{\max})}{b(\sqrt{K+1}-1)}.
\end{align*}

On the other hand, in the case of using [Exponential BS \eqref{exp_batch}], we have that
\begin{align*}
B'_K = \sum_{k=0}^{K-1}\frac{1-\beta_k}{b_k}\frac{1}{\sum_{k=0}^{K-1}\alpha_k} \leq \frac{(1-\beta_{\max})T_0E\delta}{2b_0\alpha_{\max}(\delta-1)(\sqrt{K+1}-1)}.
\end{align*}

We also have
\begin{align*}
D'_K = \sum_{k=0}^{K-1}\frac{\alpha_k^2}{1-\beta_k}\frac{1}{\sum_{k=0}^{K-1}\alpha_k} = \frac{C'_K}{1-\beta_{\max}} \leq \frac{\alpha_{\max}T\left(1+\log(K+1)\right)}{2(1-\beta_{\max})(\sqrt{K+1}-1)}.
\end{align*}

[Decaying LR] $\land$ [Increasing Beta]: We have
\begin{align*}
\alpha_k = \frac{\alpha_{\max}}{\left\lfloor\frac{k}{T}\right\rfloor+1} \geq \frac{\alpha_{\max}}{k+1} .
\end{align*}

Therefore,
\begin{align*}
\sum_{k=0}^{K-1}\alpha_k \geq \alpha_{\max}\int_{0}^{K}\frac{dk}{k+1} = \alpha_{\max}\log(K+1).
\end{align*}

Hence,
\begin{align*}
A'_K = \frac{1}{\sum_{k=0}^{K-1}\alpha_k} \leq \frac{1}{\alpha_{\max}\log(K+1)}.
\end{align*}

We also have 
\begin{align*}
\sum_{m=0}^{M-1}{\alpha'}_k^2 \leq 1 + \int_0^M\frac{dm}{(m+1)^{2}} = 1+\left(-\frac{1}{M+1}+1\right) \leq 2,
\end{align*}
which implies that
\begin{align*}
\sum_{k=0}^{K-1}\alpha_k \leq T_0\sum_{m=0}^{M-1}{\alpha'}_k^{2} \leq 2\alpha_{\max}^2T_0.
\end{align*}

Therefore,
\begin{align*}
C'_K = \frac{\sum_{k=0}^{K-1}\alpha_k^2}{\sum_{k=0}^{K-1}\alpha_k} \leq \frac{2\alpha_{\max}T_0}{\log(K+1)}.
\end{align*}

On the other hand, we obtain 
\begin{align*}
\sum_{k=0}^{K-1}(1-\beta_k) = T_0\sum_{m=0}^{M-1}\frac{1-\beta_{\min}}{(m+1)^{3/4}} \leq T_0\left(1 + \int_{0}^{M}\frac{1-\beta_{\min}}{(m+1)^{3/4}}dm\right) &= T_0\left(1+\frac{4}{3}(1-\beta_{\min})\left((M+1)^{1/4}-1\right)\right) \\
&\leq T_0\left(1+4(1-\beta_{\min})(K+1)^{1/4}\right).
\end{align*}

Therefore, in the case of using [Constant BS \eqref{constant_batch}], 
\begin{align*}
B'_K = \frac{\sum_{k=0}^{K-1}(1-\beta_k)}{b\sum_{k=0}^{K-1}\alpha_k} \leq \frac{T_0\left(1+4(1-\beta_{\min})(K+1)^{1/4}\right)}{b\alpha_{\max}\log(K+1)}.
\end{align*}

On the other hand, in the case of using [Exponential BS \eqref{exp_batch}], we have that
\begin{align*}
B'_K = \sum_{k=0}^{K-1}\frac{1-\beta_k}{b_k}\frac{1}{\sum_{k=0}^{K-1}\alpha_k} \leq \frac{(1-\beta_{\min})T_0E\delta}{b_0\alpha_{\max}(\delta-1)\log(K+1)}.
\end{align*}

We also have 
\begin{align*}
\sum_{k=0}^{K-1}\frac{\alpha_k^2}{1-\beta_k} \leq \frac{\alpha_{\max}^2T_0}{1-\beta_{\min}}\sum_{m=0}^{M-1}\frac{(m+1)^{3/4}}{(m+1)^2} &= \frac{\alpha_{\max}^2T_0}{1-\beta_{\min}}\sum_{m=0}^{M-1}\frac{1}{(m+1)^{5/4}} \leq \frac{\alpha_{\max}^2T_0}{1-\beta_{\min}}\left(1+\int_{0}^{M}\frac{dm}{(m+1)^{5/4}}\right) \\
&=\frac{\alpha_{\max}^2T_0}{1-\beta_{\min}}\left(1-\frac{1}{(M+1)^{1/4}}+1\right) \leq \frac{2\alpha_{\max}^2T_0}{1-\beta_{\min}}.
\end{align*}

Hence,
\begin{align*}
D_K' = \sum_{k=0}^{K-1}\frac{\alpha_k^2}{1-\beta_k}\frac{1}{\sum_{k=0}^{K-1}\alpha_k} \leq \frac{2\alpha_{\max}T_0}{(1-\beta_{\min})\log(K+1)}.
\end{align*}

\endproof
\end{document}